\documentclass[Afour,sagev,times]{sagej}

\makeatletter
\ExplSyntaxOn
\cs_gset:Npn \__first_footerline:
  { \hfill\null\par }
\ExplSyntaxOff
\makeatother

\usepackage[sort&compress]{natbib}

\usepackage{graphicx}
\usepackage{subcaption}
\usepackage{booktabs}
\usepackage{multirow}
\usepackage[dvipsnames,table,xcdraw]{xcolor}
\usepackage{threeparttable}
\usepackage{float}
\usepackage{array}
\usepackage{arydshln}

\usepackage{amsmath, amssymb, bm, mathtools}
\usepackage{pifont}

\usepackage{algorithm}
\usepackage{algpseudocode}

\usepackage[inline]{enumitem}
\newlist{inenum}{enumerate*}{1}
\setlist[inenum]{%
  label=(\arabic*),
  labelsep=0.4em,
  itemjoin={{; }},
  itemjoin*={{; and }},
  leftmargin=0pt,
  itemsep=0pt,
  topsep=0pt,
  parsep=0pt
}

\usepackage{hyperref}
\usepackage[nameinlink]{cleveref}
\crefname{figure}{Figure}{Figures}
\crefname{equation}{Equation}{Equations}

\usepackage[dvipsnames]{xcolor}
\usepackage{soul}

\usepackage{varwidth}

\begin{document}
\let\WriteBookmarks\relax
\def\floatpagepagefraction{1}
\def\textpagefraction{.001}

\def\volumeyear{2026}
\title{Physics-Informed Framework for Impact Identification in Aerospace Composites}

\author{Nat\'{a}lia Ribeiro Marinho\affilnum{1}, Richard Loendersloot\affilnum{1}, Jan Willem Wiegman\affilnum{2}, Frank Grooteman\affilnum{2}, and Tiedo Tinga\affilnum{1}}

\affiliation{\affilnum{1} Engineering Technology Faculty, Department of Mechanics of Solids, Surfaces  and  Systems, Dynamics Based Maintenance (DBM) Group, University of Twente (UT), Enschede, NL\\
\affilnum{2} Department of Aerospace Vehicles Integrity and Life Cycle Support (AVIL), Royal Netherlands Aerospace Centre (NLR), Marknesse, NL}

\corrauth{Nat\'{a}lia Ribeiro Marinho}

\email{n.ribeiromarinho@utwente.nl}

\begin{abstract}
Reliable impact identification in aerospace composite structures is essential for effective structural health monitoring, as low-velocity impacts may cause internal damage without visible surface indications. However, estimating impact energy from measured responses remains challenging because the inverse problem is ill-posed under sparse, noisy sensor data, nonlinear structural response, and missing excitation parameters. To address these limitations, this paper introduces a novel physics-informed impact identification (Phy-ID) framework. The proposed method integrates observational, inductive, and learning biases to combine physical knowledge with data-driven inference in a unified modelling strategy, achieving physically consistent and numerically stable impact identification. The physics-informed approach structures the input space using physics-based energy indicators, constrains admissible solutions via architectural design, and enforces governing relations via hybrid loss formulations. Together, these mechanisms limit non-physical solutions and stabilise inference under degraded measurement conditions. A disjoint inference formulation is used as a representative use case to demonstrate the framework capabilities, in which impact velocity and impactor mass are inferred through decoupled surrogate models, and impact energy is computed by enforcing kinetic energy consistency. Experimental evaluations show mean absolute percentage errors below 8\% for inferred impact velocity and impactor mass and below 10\% for impact energy. Additional analyses confirm stable performance under reduced data availability and increased measurement noise, as well as generalisation for out-of-distribution cases across pristine and damaged regimes when damaged responses are included in training. These results indicate that the systematic integration of physics-informed biases enables reliable, physically consistent, and data-efficient impact identification, highlighting the potential of the approach for practical monitoring systems.
\end{abstract}


\maketitle

\section{Introduction}
\noindent
Aerospace composite structures are highly valued for their strength-to-weight ratio, yet they remain vulnerable to undetected internal impact damage that can severely compromise structural integrity and operational safety~\citep{Dafydd_2019,yue2021damage,Giurgiutiu_2020}. To address this vulnerability, Structural Health Monitoring (SHM) systems have been developed, combining passive sensing technologies with advanced data processing algorithms to detect, locate, and infer the severity of impact events in real time. Among these tasks, estimating impact energy is crucial, as it directly correlates with damage initiation, residual strength degradation, and maintenance thresholds that guide operational decisions.

However, estimating the magnitude of an impact based on measured structural responses is a complex inverse problem that is not yet fully understood. In practical SHM implementations, the available data are often limited to sparse, and noisy measurements, with direct measurements of excitation parameters, such as impact force history, impactor mass, and impact velocity, frequently absent. As a result, the relationship between sensor signals and impact severity is non-unique, highly sensitive to noise, and influenced by factors such as structural configuration, sensor placement, and environmental conditions~\citep{Seno_2020,Seno_2021,Datta_2019}. These limitations are intensified under realistic operating conditions, where nonlinear structural behaviour, wave attenuation, and complex structural geometries further degrade numerical stability and challenge the physical credibility of inferred solutions.

Existing methodologies for estimating impact energy can be categorised into three primary groups: physics-based, data-driven, and probabilistic approaches. Physics-based approaches rely on analytical or numerical models, such as finite-element formulations~\citep{LIU2023107873,Zhou_2019,Stephen_2021}, modal representations~\citep{Ooijevaar_2015,Molina_Viedma_2021}, guided-wave propagation models~\citep{Sellami_2025,Zhang_2017}, and analytical solutions~\citep{Shariyat_2018,Zhu_2020,Alonso_2021,Bogenfeld_2018,Correas2021Analytical}, to reconstruct impact parameters from measured responses. These methods preserve physical consistency but require detailed knowledge of material properties, boundary conditions, and structural configuration, and they incur high computational cost. As a result, their applicability in real-time SHM settings is limited.

Data-driven approaches~\citep{Jung_2021,Zhu_2023,Zhong_2015,tabian2019convolutional,sharif2013smart} employ machine learning techniques to establish direct mappings from sensor measurements to impact energy, achieving efficient representations of nonlinear relationships. Their performance, however, depends strongly on the representativeness of the training data and shows sensitivity to noise and environmental variability. Moreover, the absence of explicit physical constraints limits interpretability and weakens extrapolation capability, which constrains reliability in safety-critical aerospace settings.

Probabilistic strategies merge physical modelling with data-driven inference, often incorporating uncertainty quantification to improve stability~\citep{Seno_2021,Xiao_2025,yan2017impact,ZHANG2020111882}. While these methods can lead to improved performance under controlled conditions, they frequently introduce physical information only implicitly or locally, with model classes and optimisation objectives often remaining unconstrained. As a result, improvements in predictive accuracy do not necessarily translate into physically admissible solutions, which limits interpretability and diagnostic value beyond point estimates. These issues directly affect generalisation performance and reduce confidence in outputs under realistic operational conditions, where environmental variability and measurement noise are unavoidable.

Thus, addressing the limitations identified in the current body of knowledge requires strategies that balance predictive performance with physical interpretability and robustness, ensuring that inferred impact energies remain aligned with governing relations while applicable under practical monitoring conditions. In this context, Physics-Informed Machine Learning (PIML)~\citep{karniadakis2021physics,Cross_Rogers_Pitchforth_Gibson_Zhang_Jones_2024,khalid2024advancements} offers a systematic approach by embedding physical knowledge within machine learning algorithms.

In the PIML framework, physical insight is integrated through three complementary mechanisms. First, \emph{observational bias} shapes the choice of inputs so that the data reflect physically meaningful information. Second, \emph{inductive bias} embeds prior domain knowledge assumptions directly into the model architecture, thereby restricting the range of admissible solutions. Finally, \emph{learning bias} incorporates governing relations into the optimisation objective to regularise training through explicit constitutive constraints.

Recent advances demonstrate that explicitly incorporating physical constraints into learning models improves accuracy, tolerance to operational variability and performance under distribution shifts. \citet{Rautela_2021} combined physical models and deep learning, achieving improved behaviour across excitation frequencies and noise levels compared with purely data-driven methods. Latent force models further integrate physics-based representations with probabilistic priors on unknown inputs, enabling robust estimation of structural responses under uncertain loading, as demonstrated in virtual sensing applications for offshore wind turbine structures~\citep{Zou_2023}. In addition, multi-fidelity physics-informed models improve adaptability by integrating low- and high-fidelity simulations, which enhances performance across operating regimes~\citep{Torzoni_2023}.

Despite growing interest in physics-informed learning, its systematic application to impact identification in aerospace composites is still underexplored. Existing studies often focus primarily on architectural design choices, effectively acting only on inductive bias, while physical relations are weakly enforced or applied in isolation. This limits numerical stability, interpretability, and transferability across monitoring scenarios.

This work addresses these gaps by proposing Physics-informed Impact Identification (Phy-ID), a novel, integrated physics-informed learning framework that leverages observational, inductive, and learning biases to estimate impact energy in aerospace composites. Phy-ID anchors inference to known physical relations, mitigating the ill-posedness of the inverse problem. Rather than treating impact energy as a purely statistical output, the framework infers physically meaningful parameters governing the excitation source.

To demonstrate these capabilities, the Phy-ID framework is implemented for impact identification on a relevant composite target structure. The impactor mass and impact velocity are estimated independently and combined through a known relation to obtain impact energy, for which a disjoint neural network formulation is used. The Phy-ID framework also remains adaptable to different monitoring objectives and can be extended to additional impact attributes by incorporating appropriate physical descriptors and governing relations.

The framework is evaluated using experimental impact datasets representative of operational SHM conditions, including a geometrically complex and realistic target structure, limited data availability, measurement noise, and transitions between damage states. The assessment combines predictive analytics to quantify estimation accuracy, robustness analyses with respect to data availability and noise levels, and generalisation studies addressing out-of-distribution impact scenarios. Together, these analyses provide a comprehensive evaluation of both performance and alignment with domain-knowledge principles.

\paragraph*{Contributions}
The main contributions of this work are as follows:
\begin{itemize}
\item A unified physics-informed framework, Phy-ID, that integrates observational, inductive, and learning biases to achieve physically meaningful and numerically stable impact energy estimation grounded in known physical relations.
\item The capability to inform advanced impact analyses through the inference of physically meaningful parameters, supporting event classification, source attribution, and impact characterisation.
\item Demonstration of robust and generalisable performance under representative SHM measurement conditions, including limited data availability, measurement noise, and out-of-distribution impact scenarios, supported by a comprehensive experimental evaluation conducted under controlled laboratory conditions specifically designed to emulate practical monitoring challenges.
\item Flexibility and adaptability of the framework that can be tuned to specific monitoring objectives and extended to additional impact attributes through the inclusion of relevant physical parameters and constraint formulations.
\end{itemize}

The following sections detail the integration of observational, inductive, and learning biases; present a case study illustrating the functionality of the Phy-ID framework; analyse physics-based energy indicators; and report predictive, robustness, and generalisation performance. Finally, the concluding section summarises the main findings and outlines directions for future work.

\section{General framework for Physics-Informed Impact Identification (Phy-ID)} \label{sec:phy3ID} 
\begin{figure*}
\centering
\includegraphics[width=\linewidth]{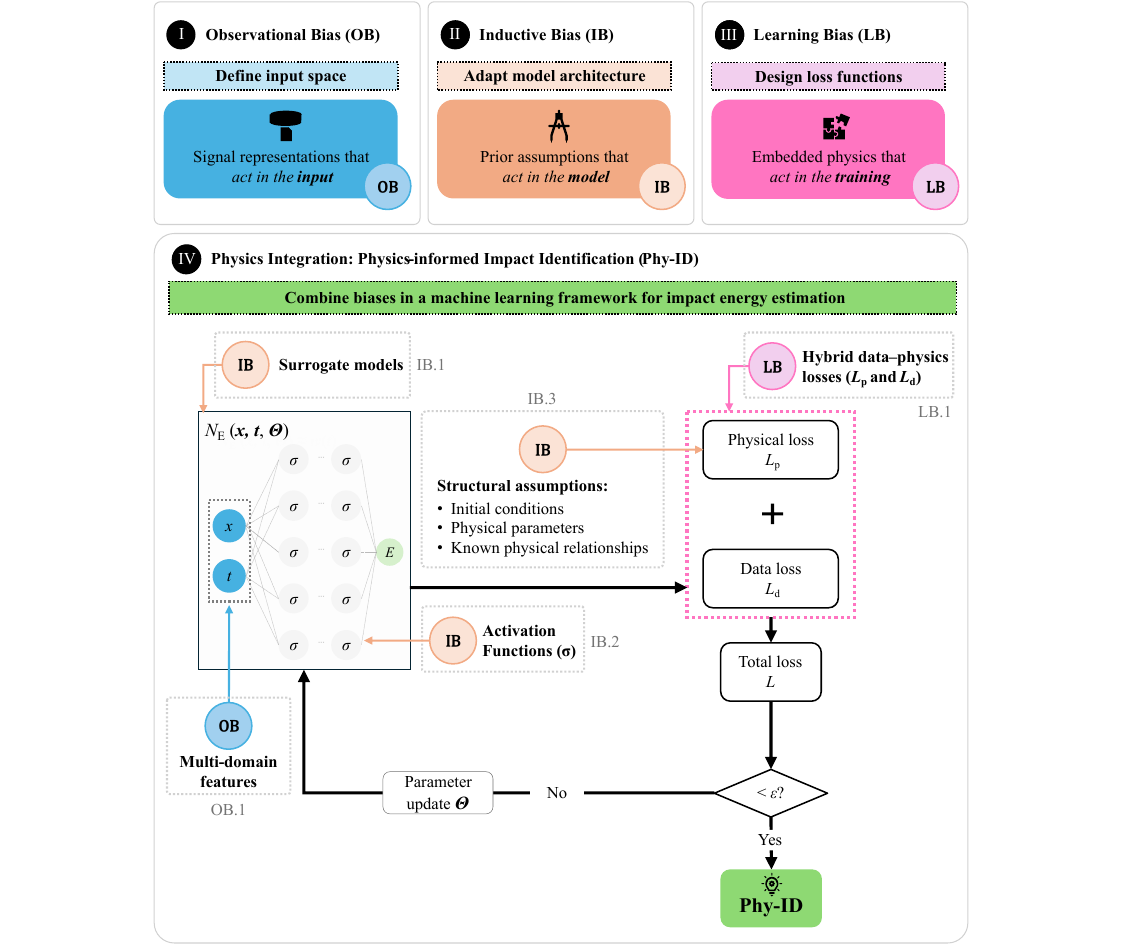}
\caption{General framework of Physics-Informed Impact Identification (Phy-ID), showing how observational, inductive, and learning biases are combined in a structured integration step to achieve physics-informed impact energy estimation.}
\label{fig:general_framework_Phy3ID}
\end{figure*}

\noindent
The proposed Phy-ID methodology applies principles of Physics-Informed Machine Learning (PIML) to the impact energy estimation task. In this approach, observational, inductive, and learning biases are embedded in a machine learning architecture to guide training towards solutions aligned with underlying physics~\citep{karniadakis2021physics}. As shown in \Cref{fig:general_framework_Phy3ID}, the framework consists of four parts: Observational Bias (Part I), Inductive Bias (Part II), Learning Bias (Part III), and Physics Integration (Part IV).

In Part I, observational bias defines how the physical system is represented through the available data. It governs the selection, organisation, and synthesis of input information to ensure that the model observes realistic structural behaviour through signal representations that capture a comprehensive description of the system state.

In Part II, inductive bias governs how prior knowledge or constraints shape the model architecture. It embeds assumptions that determine how information is processed within the network structure. Broadly, this bias defines how inputs and outputs relate through physically-motivated mappings.

In Part III, learning bias regulates the optimisation process by enforcing agreement between model predictions and governing physical principles via the design of the loss function. It defines the learning objectives that guide convergence towards solutions consistent with both empirical observations and underlying physical laws during training.

In Part IV, physics integration operationalises the Phy-ID method by combining all three forms of bias within a unified framework for impact energy estimation. This stage characterises the method by integrating data, model structure, and training objectives within a single optimisation process. Within this integration step, the three forms of bias interact coherently to shape the model and limit the emergence of non-physical predictions that may otherwise satisfy purely data-driven objectives.

First, observational bias is introduced through the way physical information enters the learning process via the input space. At this stage, measured data are transformed into multi-domain features (OB.1 in \Cref{fig:general_framework_Phy3ID}) that describe the structural response in time, frequency, and time-frequency domains. These features act as physics-based energy indicators that capture key information on the effect of impact loading in the measured signals. Although often overlooked in machine learning studies, a well-designed observational bias provides an optimised and efficient input space, leading to measurable improvements in prediction accuracy and interpretability, as demonstrated in~\citet{marinho2025energyindicators}.

Subsequently, the inductive bias incorporates prior physical knowledge into the model architecture. It is implemented through surrogate models (IB.1 in \Cref{fig:general_framework_Phy3ID}), appropriate activation functions (IB.2 in \Cref{fig:general_framework_Phy3ID}), and structural assumptions informed by the system behaviour (IB.3 in \Cref{fig:general_framework_Phy3ID}). In the current implementation, surrogate models within Neural Network (NN) architectures infer relevant parameters that characterise the impact event, maintaining consistency through known physical relationships. Realistic initial conditions constrain the model response at the onset of impact, while smooth activation functions ensure stable differentiation during training, thereby improving numerical stability and reliable gradient computation.

Finally, the learning bias governs the optimisation process that drives model training. It operates through a hybrid loss function (LB.1 in \Cref{fig:general_framework_Phy3ID}) that merges data-based and physics-motivated terms. The data term enforces agreement between predictions and experimental measurements, while the physics term constrains the solution to satisfy governing physical laws. Model parameters are iteratively updated until convergence, balancing agreement with measured data and imposed physical relations.

Although presented here for impact energy estimation, the proposed framework remains adaptable to other monitoring objectives and application domains. Additional physical knowledge or alternative model formulations can be incorporated to estimate different physical quantities or to support subsequent stages of structural health monitoring.


\section{Implementation of the Phy-ID framework} \label{sec:disjoint_pinn}
\noindent
This section presents an implementation strategy of the Phy-ID framework designed to demonstrate its capabilities rather than to define a unique or prescriptive solution. The implementation serves as a concrete example of how observational, inductive, and learning biases can be combined within a coherent formulation to support physically meaningful inference. The focus extends beyond prediction accuracy, showing how governing physical relations can guide training and enable the estimation of latent physical parameters associated with the excitation process.

In this implementation, the kinetic energy relation provides a direct link between the measured responses and the parameters inferred by the model. Its analytical form is compact and well established in mechanics, yet sufficiently expressive to structure the training process around a physically grounded constraint. In classical mechanics, the kinetic energy of a rigid impactor is given by
\begin{equation} \label{kinetic_energy}
E = \tfrac{1}{2} m v_0^2 ,
\end{equation}
where \(m\) is the impactor mass and \(v_0\) is the impact velocity at the instant of contact. Enforcing this relation during optimisation constrains the predictions to admissible combinations of mass and velocity and ensures consistency with the measured impact energy.

Building on this structural prior, the impactor mass and impact velocity are defined as the target outputs for inferring impact energy and characterising the impact event. This choice is deliberate. Although impact energy could be predicted directly using a single surrogate model, such a formulation would treat energy as an isolated target and would not require the model to recover physically interpretable combinations of mass and velocity. By instead identifying mass and velocity and combining them through the kinetic energy relation, the formulation preserves the physical structure of the impact process and ensures mechanical consistency between the inferred parameters and the resulting energy estimate. Estimating these quantities also enriches the analysis of the impact event by providing more information on whether damage may have formed and what type of damage is likely.

However, estimating both parameters within a single multi-output network leads to unbalanced optimisation because they influence the measured impact response in fundamentally different ways. Velocity dominates the optimisation, as it governs the local and transient dynamics of the impact event. It controls the amplitude, duration, and frequency content of the induced stress waves and strongly affects measurable signal features such as peak amplitude, rise time, and dominant frequency, which are highly sensitive to variations in impact energy~\citep{marinho2025energyindicators,Andrew_2019,Zhang_2024}. 

In contrast, the effect of mass is more diffuse, influencing the overall energy transfer and global dynamic properties. Its effect on the stress waves is weaker and often masked, especially when sensor arrays are sparse~\citep{Aryal_2019,Artero_Guerrero_2015,Mukhopadhyay_2020}. Consequently, information about mass appears in the measured signals with lower observability and in a more indirect manner than that of velocity. During joint training, the optimisation therefore tends to prioritise velocity learning and underestimates mass.

Moreover, the kinetic energy relation can further accentuate this effect because velocity enters the expression quadratically. Although this is not the primary source of the imbalance optimisation behaviour described above, it increases the sensitivity of the energy-based loss to velocity errors relative to mass errors, reinforcing the dominance of velocity during joint optimisation. To mitigate this imbalance, the Phy-ID framework utilises a disjoint PINN that sequentially estimates mass and velocity using decoupled networks, allowing each parameter to be learned in a more balanced manner, thereby improving optimisation stability and maintaining physical consistency.

Using a disjoint architecture, the problem is formulated as a sequential training algorithm. The concept of stabilising training through sequential optimisation of separate networks has been explored in other contexts, such as the thermochemical curing of composite-tool systems~\citep{AMININIAKI2021113959}. Although the application, governing equations, and basis formulation in \citet{AMININIAKI2021113959} differ fundamentally from the present work, the principle of decoupled networks trained in sequence motivates the architectural choice adopted here.

Accordingly, the Phy-ID method uses two decoupled surrogate models: a displacement network $w_\theta$ and a mass network $m_\phi$. Both are implemented as Fully Connected Neural Networks (FCNNs). The input feature matrix $\boldsymbol{x} \in \mathbb{R}^{N\times F}$ represents $N$ impact events described by $F$ features defining the input space. The displacement surrogate model $w_\theta: \mathbb{R}^{N\times F} \times \mathbb{R} \to \mathbb{R}$ maps $\boldsymbol{x}$ and a time vector $t$ to the transverse displacement $w_\theta(\boldsymbol{x},t)$. The mass surrogate model $m_\phi: \mathbb{R}^{N\times F} \to \mathbb{R}_{>0}$ operates on the same feature matrix to predict a strictly positive mass $m_\phi(\boldsymbol{x})$ through a softplus output layer with a small offset $\varepsilon=10^{-6}$. 

Enforcing positive mass ensures physical consistency, as negative or zero mass would lead to non-physical, impossible conditions, including undefined or negative kinetic energy. In contrast, displacement may assume positive or negative values depending on the direction of motion relative to a reference axis. This variability reflects real structural behaviour and does not compromise physical consistency, because the model does not rely on the chosen reference frame.

The displacement output is then used to infer the impact velocity, which is obtained by differentiating the displacement with respect to time and evaluating it at $t=0$,
\begin{equation}
v_0(\boldsymbol{x}) = \left.\frac{\partial w_\theta(\boldsymbol{x},t)}{\partial t}\right|_{t=0}.
\end{equation}

Embedding impact velocity as the initial time derivative of a displacement field anchors the formulation in structural dynamics, where displacement and its first derivative define the system state at the instant of contact. This representation also allows the physically meaningful initial condition to be enforced during training through the loss formulation introduced later in this section. In addition, by parameterising both the zero-displacement condition and the initial velocity within the same surrogate field, the formulation avoids introducing additional independent quantities and improves numerical conditioning. Because the derivative is obtained through automatic differentiation, incorporating this relation introduces only minimal additional complexity. As a result, the learning problem remains physically grounded and numerically stable while retaining the capacity to accommodate additional governing relations if required.

Because the impact velocity enters the kinetic energy expression as a squared term, its sign does not affect the energy calculation; therefore, enforcing positivity is unnecessary. Together, the predicted mass and inferred velocity are linked through the kinetic energy relation in \Cref{kinetic_energy}, ensuring physical compatibility between the two surrogate networks. 

\begin{figure*}
\centering
\includegraphics[width=0.75\linewidth]{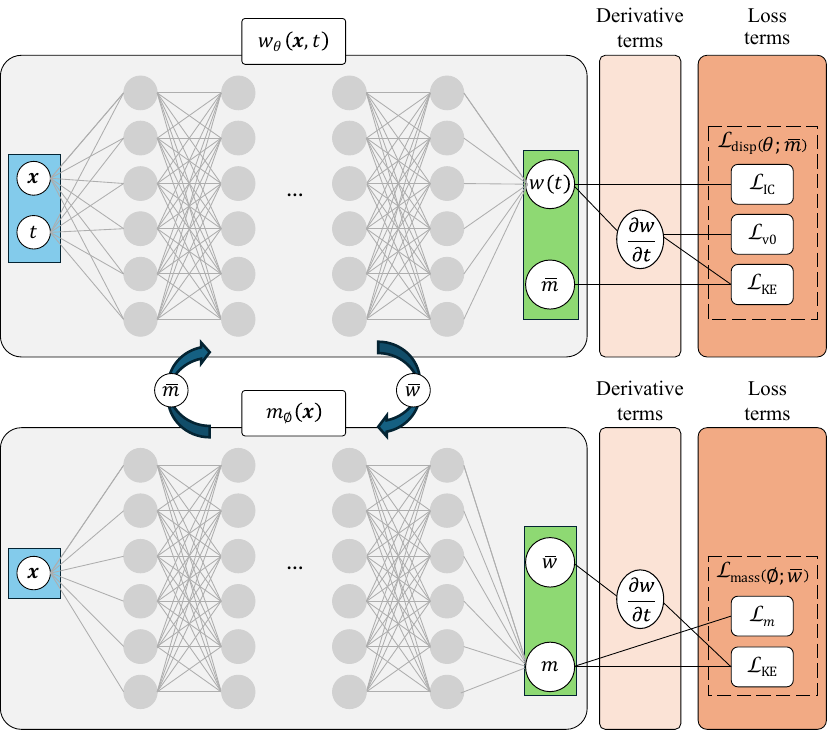}
\caption{High-level overview of the sequential disjoint training process: the displacement network (top) generates a velocity estimate for the impactor mass network (bottom), which in turn produces a mass estimate fed back to the displacement model.}
\label{fig:sequential_training}
\end{figure*}

This coupling forms the basis of the sequential training process summarised in \Cref{fig:sequential_training}, which provides a high-level overview of how the two networks exchange information during alternating optimisation. At each iteration, the parameters of the mass model are fixed, and the displacement network $w_\theta$ is optimised with respect to its loss function $\mathcal{L}_{\mathrm{disp}}$. In the first iteration, the measured mass $m_{\text{obs}}$ is used to initialise the displacement model. Once updated, the displacement parameters are held constant, and the mass network $m_\phi$ is optimised with respect to its loss $\mathcal{L}_{\mathrm{mass}}$. This alternating process continues until convergence, ensuring balanced accuracy and adherence to underlying physical constraints between the coupled quantities.

The loss functions for both $\mathcal{L}_{\mathrm{disp}}$ and $\mathcal{L}_{\mathrm{mass}}$ combines measurement-based and physics-based components, corresponding to the data loss ($L_d$) and physics loss ($L_p$) illustrated in \Cref{fig:general_framework_Phy3ID}. In general form, hybrid losses can therefore be expressed as
\begin{equation}
\mathcal{L} = \mathcal{L}_d + \mathcal{L}_p.
\end{equation}
The displacement loss is defined as
\begin{equation}
\mathcal{L}_{\text{disp}}
=
\underbrace{\lambda_{v0}\mathcal{L}_{v0}}_{L_d}
+
\underbrace{\lambda_{\text{IC}}\mathcal{L}_{\text{IC}}}_{L_p}+\underbrace{\lambda_{\text{KE}}\mathcal{L}_{\text{KE}}}_{L_p\text{, }L_d},
\end{equation}
where the weighting factors $\lambda_{v0}$, $\lambda_{\text{IC}}$ and $\lambda_{\text{KE}}$ control the relative contribution of the velocity, initial condition, and kinetic energy terms in this phase. The individual components are
\begin{equation}
    \begin{split}
        \mathcal{L}_{v0} &= \lVert v_{0,\text{obs}} - v_0(\boldsymbol{x}) \rVert^2\\[4pt]
        \mathcal{L}_{IC} &= \lVert w_\theta(\boldsymbol{x},0) \rVert^2\\[4pt]
        \mathcal{L}_{KE} &= \lVert E_{\text{meas}} - \tfrac{1}{2}\,\overline{m}\,v_0^2(\boldsymbol{x}) \rVert^2.
    \end{split}
\end{equation}
Here, $\overline{m}$ denotes the fixed mass (either the measured $m_{\text{obs}}$ in the first iteration or the predicted mass in later iterations). These terms penalise, respectively, deviations from the observed impact velocity $v_{0,\text{obs}}$, violations of the zero-displacement initial condition (IC), and inconsistencies with kinetic energy (KE).

The mass network loss follows an analogous form,
\begin{equation}
\mathcal{L}_{\text{mass}}
=
\underbrace{\lambda_{m}\mathcal{L}_{m}}_{L_d}
+
\underbrace{\lambda_{\text{KE},m}\mathcal{L},_{\text{KE},m}}_{L_p\text{, }L_d},
\end{equation}
with
\begin{equation}
    \begin{split}
        \mathcal{L}_{m} &= \lVert m_{\text{obs}} - m_\phi(\boldsymbol{x}) \rVert^2\\[4pt]
        \mathcal{L}_{\text{KE},m} &= \lVert E_{\text{meas}} - \tfrac{1}{2} m_\phi(\boldsymbol{x}) \overline{v}_0^2(\boldsymbol{x}) \rVert^2 \Big. \text{.}
    \end{split}
\end{equation}
The first term enforces agreement between predicted and measured mass ($m_\phi$ and $m_\text{obs}$), while the second ensures consistency between the predicted mass $m_\phi$, fixed velocity $\overline{v}_0$ resulting from the displacement NN, and observed energy $E_\text{meas}$. The factors $\lambda_m$ and $\lambda_{\text{KE},m}$ act as the corresponding weighting coefficients for the mass phase.

The weighting factors introduced above regulate the influence of measurement terms and physics-based terms during training, preventing dominance by any single component. For this study, the values are assigned manually, a common and effective approach for demonstration purposes that provided stable training and adequate performance. Alternative schemes, such as adaptive weighting~\citep{Wang_2021} or uncertainty-based scaling~\citep{Perez_2023, Xiang_2022}, may benefit more complex or sensitive applications and could be explored in future work.

Building on the loss formulations, the two optimisation objectives summarise the sequential training process,
\begin{equation}
    \begin{split}
    \theta^\ast=\arg\min_\theta \mathcal{L}_{\text{disp}}(\theta;\overline{m}) \quad \text{and} \\[4pt]
    \phi^\ast=\arg\min_\phi \mathcal{L}_{\text{mass}}(\phi;\overline{v}_0),
    \end{split}
\end{equation}
where $\theta$ and $\phi$ are updated alternately according to the displacement- and mass-phase to ensure balanced accuracy and physical consistency. Algorithm~\ref{alg:phy3id} provides a detailed description of the sequential training steps. The algorithm takes as input a feature matrix $\boldsymbol{x}\in\mathbb{R}^{N\times F}$, measured energy $E_{\mathrm{meas}}\in\mathbb{R}^{N\times 1}$, measured mass $m_{\mathrm{obs}}\in\mathbb{R}^{N\times 1}$, and measured impact velocity $v_{0,\mathrm{obs}}\in\mathbb{R}^{N\times 1}$, corresponding to $N$ impact events. The final outputs $\hat{m}$ and $\hat{v}_0$ are combined through the classical kinetic energy relation introduced in ~\Cref{kinetic_energy} to predict the impact energy. 
\begin{algorithm}
\small 
\caption{\\ Physics-Informed Impact Identification (Phy-ID) \\ Sequential training} \label{alg:phy3id} \begin{algorithmic}[1]
\State \textbf{Input:} $\boldsymbol{x},E_{\text{meas}},m_{\text{obs}},v_{0,\text{obs}},t$. \State \textbf{Models:} $w_\theta(\boldsymbol{x},t)$ (displacement FCNN), $m_\phi(\boldsymbol{x})$ (mass FCNN). \For{cycle $c=0,\dots,C-1$} \State \textbf{Phase A: Displacement optimisation} 
\State Fix $\overline{m}=m_{\text{obs}}$ if $c=0$, else $\overline{m}=m_\phi(\boldsymbol{x})$. 
\State Solve $\theta^\ast=\arg\min_\theta \mathcal{L}_{\text{disp}}(\theta;\overline{m})$ with Adam optimiser. \State \textbf{Phase B: Mass optimisation} 
\State Compute $\overline{v}_0$ from fixed $w_\theta$. 
\State Solve $\phi^\ast=\arg\min_\phi \mathcal{L}_{\text{mass}}(\phi;\overline{v}_0)$ with Adam optimiser. \EndFor 
\State \textbf{Output:} $\hat m,\hat v_0$. 
\vspace{1pt}
\Statex \textit{Notation:} $\overline{(\cdot)}$ denotes a fixed value; $(\cdot)^\ast$ marks the optimal solution; and $\hat{(\cdot)}$ indicates predicted quantities.
\end{algorithmic} 
\end{algorithm}

\begin{table*}
    \centering
    \caption{Hyperparameter search space explored for the displacement and mass FCNNs.}
    \label{tab:grid_search_phy3id}
    \begin{tabular}{@{}lcc@{}}
        \toprule
        \textbf{Parameter} 
            & \textbf{Displacement network, $w_\theta(\boldsymbol{x},t)$} 
            & \textbf{Mass network, $m_\phi(\boldsymbol{x})$} \\
        \midrule
        fully connected layer size, $n_h$ 
            & 32, 64, 128, 256 
            & 16, 32, 64 \\
        number of hidden layers, $L_h$
            & 2, 3, 4 
            & 2, 3 \\
        learning rate, $lr$ 
            & $10^{-2}$, $10^{-3}$ 
            & $10^{-2}$, $10^{-3}$ \\
        activation function 
            & Sigmoid, Softplus, Tanh
            & Sigmoid, Softplus, Tanh\\
        \bottomrule
    \end{tabular} 
    \vspace{1em}
\end{table*}

It is important to distinguish between training and inference phases. During training, the model receives both input features and the measured physical quantities (measured energy $E_{\mathrm{meas}}$, measured mass $m_{\mathrm{obs}}$, and measured impact velocity $v_{0,\mathrm{obs}}$), which act as supervised targets for optimising the surrogate models through the hybrid loss. During inference, the trained models take only the feature matrix 
$\boldsymbol{x}$ as input. The learned model is then used to estimate the physical parameters and compute the impact energy from the predicted mass $\hat{m}$ and impact velocity $\hat{v}_0$.
\begin{table*}
\centering 
\caption{Implementation settings for the disjoint PINN architecture.} 
\label{tab:phy3id_hyperparams} 
\begin{tabular}{ll} 
\hline \textbf{Parameter} & \textbf{Value} \\ \hline \multicolumn{2}{l}{\textbf{Displacement network $w_\theta(\boldsymbol{x},t)$}} \\ architecture & FCNN, 3 hidden layers of size 256, activation $\tanh$ \\ output & impact velocity, $\hat v_0$ \\ learning rate & $1\times 10^{-2}$ \\ loss function & $\mathcal{L}_{\text{disp}}=\lambda_{v0}\mathcal{L}_{v0}+\lambda_{\text{ic}}\mathcal{L}_{IC}+\lambda_{\text{ke}}\mathcal{L}_{KE}$ \\ weighting factors & $\lambda_{v0}=1\times 10^{-4}$, $\lambda_{\text{IC}}=1\times 10^{-6}$ and $\lambda_{\text{KE}}=1\times 10^{-4}$ \\ \arrayrulecolor{gray!50} \midrule \multicolumn{2}{l}{\textbf{Mass network $m_\phi(\boldsymbol{x})$}} \\ architecture & FCNN, 3 hidden layers of size 64, activation softplus \\ output & impactor mass, $\hat m$ \\ learning rate & $1\times 10^{-3}$ \\ loss function & $\mathcal{L}_{\text{mass}}=\lambda_m\mathcal{L}_m+\lambda_{\text{ke},m}\mathcal{L}_{KE,m}$ \\ weighting factors & $\lambda_m=1\times 10^{-6}$ and $\lambda_{\text{KE},m}=1\times 10^{-4}$ \\   \hline \multicolumn{2}{l}{\textbf{Training procedure}} \\ epochs & max 10000 per phase \\ outer cycles & max $C=10$  \\ batch size & full-batch \\ optimiser & Adam  \\
criterion & MSELoss \\
patience & 500 \\
\arrayrulecolor{black} \bottomrule \end{tabular} \vspace{1em}
\end{table*} 

The implementation presented in this section operationalises the bias structure introduced in~\nameref{sec:phy3ID}. The observational bias (OB.1) is realised through the feature matrix, which contains the multi-domain signal representations describing the measured structural response. The inductive biases are embedded in the model formulation: IB.1 corresponds to the surrogate neural network models used to represent displacement and mass; IB.2 arises from the use of smooth activation functions that enable stable differentiation and consistent parameter estimation while aligning with the expected behaviour of the physical fields; and IB.3 is introduced through structural constraints such as the kinetic relation between displacement and velocity, the initial condition, the positivity constraint on mass, and the kinetic energy relation. Finally, the learning bias (LB.1) is implemented through the hybrid loss functions and the sequential optimisation procedure, which combine measurement-based terms and physics-based constraints to guide training towards predictions grounded in structural mechanics.

\subsection{Architecture details}
\label{architecture_NN}
\noindent
The architecture for the disjoint PINN supports the sequential training strategy described earlier in this section. It consists of two separate Fully Connected Neural Networks (FCNNs), each dedicated to a specific regression task: the displacement network estimates the impact velocity, and the mass network predicts the impactor mass. Although the subnetworks are trained separately, they share the same feature matrix as input, and their outputs are coupled in the loss function via a kinetic-energy relation to recover the physical quantities required for energy prediction.

To select the network architecture, a grid search was performed using five-fold cross-validation. For each fold, the dataset was split into 80\% for training and 20\% for testing. The search tested different combinations of hidden layer numbers ($L_h$), layer sizes ($n_h$), learning rates ($lr$), and activation functions ($\sigma$) for both surrogate models. The search space for both networks is summarised in \Cref{tab:grid_search_phy3id}. The activation functions tanh (hyperbolic tangent), softplus, and sigmoid introduce inductive biases into the network architecture. Their continuous and smooth behaviour supports stable differentiation, which is required to evaluate the derivatives used in the physics-based constraints. In addition, the softplus output layer ensures that the predicted mass is non-negative. These choices, therefore, mainly control the regularity and boundedness of the learned representations while remaining compatible with the physical formulation.

\begin{figure*}
  \centering
  \begin{subfigure}{\textwidth}
    \centering
    \includegraphics[width=\linewidth]{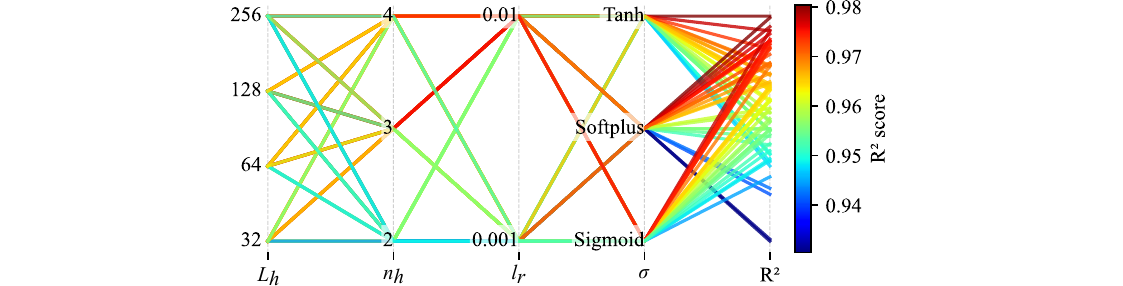}
    \caption{\centering Displacement network $w_\theta(\boldsymbol{x},t)$.}
    \label{fig:parallel_coordinates_disp}
  \end{subfigure}
  \begin{subfigure}{\textwidth}
    \centering
    \includegraphics[width=\linewidth]{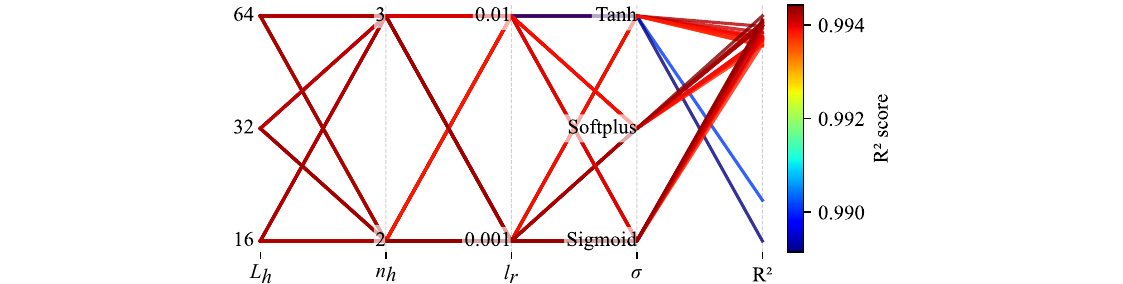}
    \caption{\centering Mass network $m_\phi(\boldsymbol{x})$.}
    \label{fig:parallel_coordinates_mass}
  \end{subfigure}
  \caption{Parallel coordinate plots showing the performance of the displacement and mass networks across hyperparameter combinations. ($n_h$: fully connected layer size; $L_h$: number of fully connected layers; \textit{lr}: learning rate; and $\sigma$: activation functions.)}
  \label{fig:parallel_coordinates}
\end{figure*}

The performance of each architectural configuration was assessed using the mean coefficient of determination $R^2$ across the test folds, with the corresponding validation trends shown in \Cref{fig:parallel_coordinates}. The colour scale shows the average $R^2$ for each parameter combination and highlights the most effective settings within the explored search space. Guided by these trends, a final set of hyperparameters was selected. These values, together with the remaining implementation details used in the sequential scheme, are listed in \Cref{tab:phy3id_hyperparams}, and this configuration was adopted for all subsequent analyses.

\subsection{Dataset}
\label{sec:dataset}
\begin{figure*}
\centering
  \begin{subfigure}[t]{0.14\textwidth}
      \includegraphics[width=\linewidth]{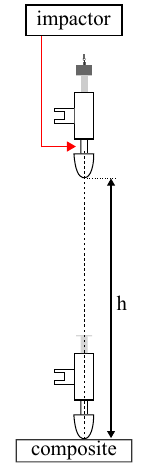}
      \caption{\centering}
      \label{fig:dt_assembly}
  \end{subfigure}
  \qquad
  \begin{subfigure}[t]{0.28\textwidth}
      \includegraphics[width=\linewidth]{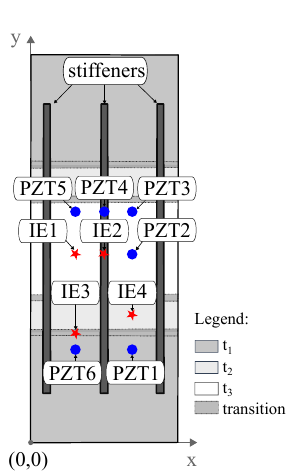}
      \caption{\centering}
      \label{fig:experimentaldesign_element}
  \end{subfigure}
  \qquad
  \begin{subfigure}[t]{0.4\textwidth}
      \includegraphics[width=\linewidth]{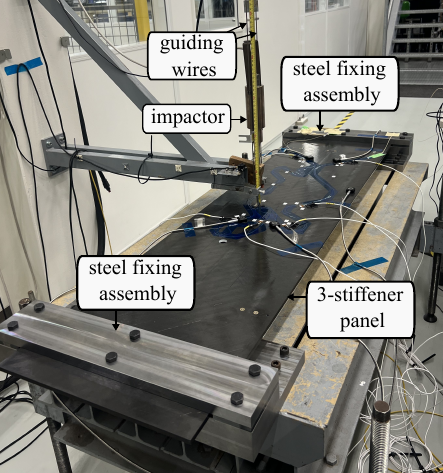}
      \caption{\centering}
      \label{fig:setup_element}
  \end{subfigure}

  \caption{Schematic impact test configuration; 
  \textcolor{blue}{$\bullet$} piezoelectric sensors (PZT) and 
  \textcolor{red}{$\star$} Impact Locations (IE): 
  (a) drop-tower assembly with impact height ($h$) set by impact energy; 
  (b) stiffened thermoplastic composite structure (\(1600 \times 370\)~mm) 
  with varying thicknesses $t_1 = 8.26$~mm, $t_2 = 6.30$~mm, and $t_3 = 5.46$~mm; (c) experimental set-up, with stiffeners at the bottom-side of the panel.
  Adapted from~\citep{marinho2025evaluating,marinho2024impact}.}
  \label{fig:subfig_setup} 
\end{figure*}

\noindent
The proposed method is validated using experimental data from intermediate-mass~\citep{olsson2000mass} impact tests on a stiffened thermoplastic composite structure. This element-level configuration~\citep{faa2009ac} is geometrically complex, offering a realistic setting for assessing the method. The composite structure features three stiffeners and three distinct thickness regions, with a quasi-isostatic lay-up, and nominal dimensions of \(1600~\text{mm} \times 370~\text{mm}\). The schematic of the test configuration is illustrated in \Cref{fig:subfig_setup}, while the impact locations and sensor coordinates are summarised in \Cref{setup_coordinates}. 
\begin{table}
  \centering
  \caption{Impact locations and sensor network coordinates.}
    \begin{tabular}{p{12mm}cc}
    \multirow{2}[0]{*}{\textbf{ID}} & \multicolumn{2}{p{27.5mm}}{\textbf{Coordinates [mm]}} \\
            & \textit{x} & \textit{y}       \\    \hline
     PZT1  & 255   & 428 \\
     PZT2  & 252   & 818 \\
    PZT3  & 250   & 945 \\
     PZT4  & 187   & 945 \\
     PZT5  & 125   & 945 \\
     PZT6  & 125   & 427 \\
   IE1   & 124   & 816 \\
    IE2   & 182   & 818 \\
    IE3   & 123   & 491 \\
    IE4   & 252   & 559 

    \end{tabular}
    \label{coord_element}
  \label{setup_coordinates}
\end{table} 

Impact tests follow the ASTM D7136M-15~\citep{ASTMD7136} standard, ensuring consistent loading and measurement conditions throughout the experimental campaign. Drop-weight impacts were applied on the outboard side of the composite component. Two hemispherical impactors, each with a diameter of 16~mm but differing in mass of 2.238~kg, 2.356~kg and 5.510~kg, delivered the impacts. A network of six piezoelectric (PZT) sensors recorded the impact responses. Additional details on the material system, sensor configuration, and data-collection procedure are available in \citet{marinho2024impact}. The resulting dataset comprises 73 distinct impact events, with energies ranging from 3.74 J to 80.95 J, including both pristine and damaged conditions. An overview of the impact energy distribution statistics is presented in \Cref{tab:numstats}.
\begin{table}
    \caption{Impact energy statistics.}
    \centering
    \begin{tabular}{lc}
    \toprule
    Number of unique targets                & 73  \\
    Minimum energy [J]                      & 3.74 \\
    Maximum energy [J]                      & 80.95 \\
    Mean energy [J]                         & 35.92 \\
    Median energy [J]                       & 29.83 \\
    Standard deviation [J]                  & 24.04 \\
    \bottomrule
    \end{tabular}
    \label{tab:numstats}
\end{table}

Phased-array ultrasonic inspections were performed after every impact event to identify the energy level associated with the damage onset. The measurements were performed using an Olympus Omniscan M-PA16-128 system with a phased-array transducer containing 128 elements, sampled at 100 MHz, operating with 256 focal laws and an 80 V pulse voltage per element, with automatic probe recognition. A water-based couplant assured proper signal transmission. The scans provide complementary views, including backwall attenuation, reflection maps, and cross-sectional images, that reveal the progression of subsurface damage.
\begin{figure*}
\centering
  \begin{subfigure}[t]{0.69\textwidth}
      \includegraphics[width=\linewidth]{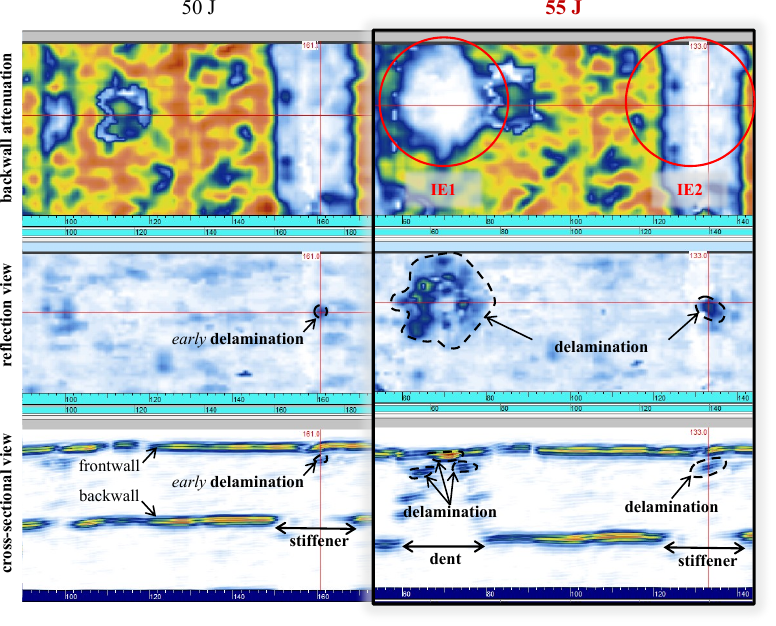}
      \caption{\centering Impact locations IE1 and IE2.}
      \label{damageonset_IE1_IE2}
  \end{subfigure}
  \qquad
  \begin{subfigure}[t]{0.3\textwidth}
      \includegraphics[width=\linewidth]{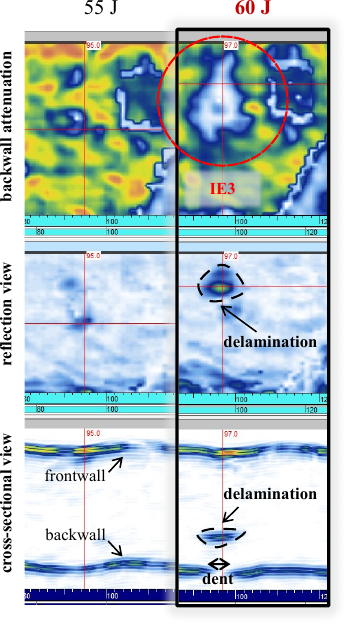}
      \caption{\centering Impact location IE3.}
      \label{damageonset_IE3}
  \end{subfigure}
  \qquad \hspace{2em}
  \begin{subfigure}[t]{0.3\textwidth}
      \includegraphics[width=\linewidth]{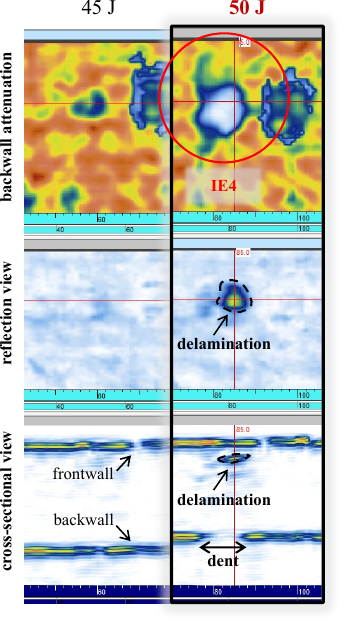}
      \caption{\centering Impact location IE4.}
      \label{damageonset_IE4}
  \end{subfigure}
    \caption{
    Phased array ultrasonic scans for different impact locations acquired after the impact events that precede and produce damage, highlighting the impact energy associated with damage onset. Each scan shows backwall attenuation, reflection view and cross-sectional view.}
  \label{damageonset}
\end{figure*}

As shown in \Cref{damageonset}, the ultrasonic response captured damage initiation under progressive loading conditions. Analysis at \textit{IE1} and \textit{IE2} (\Cref{damageonset_IE1_IE2}) showed that impacts of 50~J resulted in only minor changes in the backwall signal, with subtle irregularities more pronounced at \textit{IE2}. However, at 55~J, scanning revealed significant delamination and surface dents, indicating the onset of damage. Further observations at \textit{IE3} (\Cref{damageonset_IE3}) indicated limited variation at 55 J, while the scan at 60~J displayed distinct delamination patterns and localised surface marks. For \textit{IE4} (\Cref{damageonset_IE4}), views at 45~J exhibited no signs of damage, while the 50~J inspection revealed visible delamination and surface dents. These results suggest an average damage onset energy of approximately 55~J for the stiffened composite panel. This threshold is used to define and interpret various impact scenarios and data availability conditions in the following analyses.

\subsection{Data processing}
\noindent
To systematically evaluate the proposed method, a structured data processing workflow has been implemented, focusing on three key objectives: predictive performance, robustness against limited and noisy data, and generalisation to unseen impact situations. While all modelling choices remain fixed across objectives, the optional processing steps are tailored to address data conditioning requirements for each analysis.

\begin{figure*}
\centering
\includegraphics[width=\linewidth]{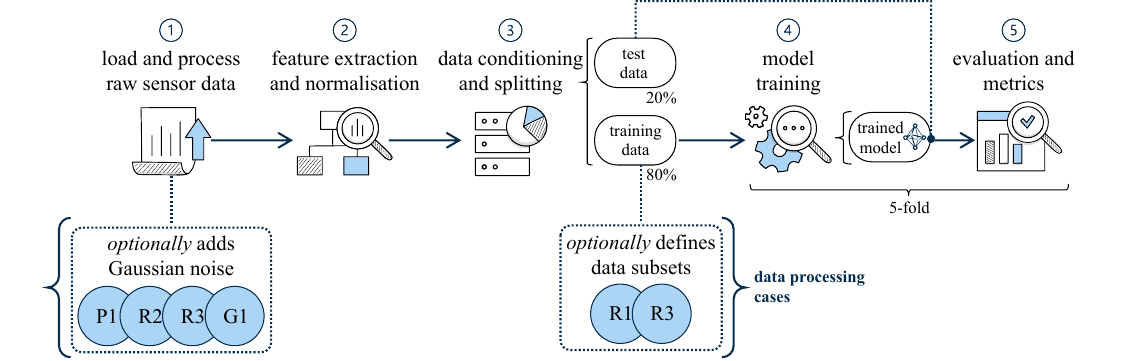}
\caption{Data-processing workflow highlighting the associated analysis cases: P1 (predictive analytics), R1 (data availability sweep), R2 (noise sweep), R3 (combined availability and noise), and G1 (out-of-distribution training ranges).}
\label{fig:DP_workflow}
\end{figure*}

A detailed summary of the examined data processing cases is presented in Table \ref{tab:data_processing_cases}. This table outlines the different scenarios for predictive analytics (P1), robustness analyses (R1–R3), and generalisation assessments (G1), each characterised by variations in available impact data, noise levels, and the nominal impact energy range employed during training. The general data processing workflow is illustrated in Figure \ref{fig:DP_workflow}, highlighting where case-specific adaptations apply.
\begin{table*}
\caption{Summary of data processing cases for predictive analytics, robustness analysis and generalisation assessment.}
\centering
\small
\begin{tabular}{l p{45mm} cc c c}
\toprule
\multirow{2}{*}{\textbf{Case\textsuperscript{a,b}}}  & \multirow{2}{=}{\textbf{Description}}  & \multicolumn{2}{c}{\textbf{Impact data}}  & \textbf{Training scenario}  & \multirow{2}{*}{\textbf{Samples}} \\
\cmidrule(lr){3-4} &  & \textbf{Data avail. [\%]} & \textbf{Noise [\%]} 
& \textbf{Energy range\textsuperscript{c} [J]} &  \\
\midrule
\textbf{P1}  & Predictive analytics & 100 & 1.0  & 4--80  & 146 \\
\midrule
\textbf{R1}  & Data availability sweep  & 25--100 & 0.0  & 4--80  & variable \\
\arrayrulecolor{gray!50}
\midrule
\textbf{R2} & Noise sweep  & 100 & 0--5  & 4--80  & 73 \\
\midrule
\textbf{R3}  & Combined availability and noise & 25--100 & 0--5  & 4--80  & variable \\
\arrayrulecolor{black}
\midrule
\textbf{G1}  & Out-of-distribution training ranges  & 100 & 1.0  & variable  & 146 \\
\bottomrule
\end{tabular}
\vspace{0.6em}
\begin{tablenotes}
\small
\item \textsuperscript{a} All cases use the same input feature set.
\item \textsuperscript{b} All models use the same FCNN architecture for displacement and mass subnetworks (see \nameref{architecture_NN}).
\item \textsuperscript{c} Energy limits refer to nominal impact energy, which may differ slightly from measured values recorded by the drop tower instrumentation.
\end{tablenotes}
\label{tab:data_processing_cases}
\end{table*}
\begin{figure}
\centering
\includegraphics[width=\linewidth]{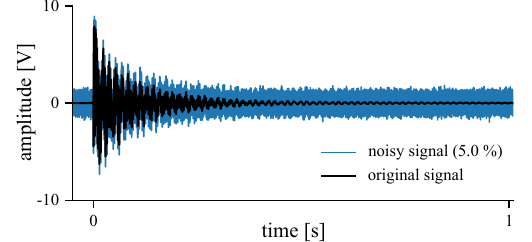}
\caption{Comparison between the original and noisy signals recorded by PZT5 due to an impact event of 10~J at impact location IE1. \\
\footnotesize  PZT: piezoelectric}
\label{fig:example_noise}
\end{figure}

All cases follow the same core processing pipeline. Initially, raw impact responses are processed to yield time signals of consistent quality, as previously outlined by \citet{marinho2025evaluating}. Feature extraction subsequently transforms each processed response into a fixed-length vector, forming the feature matrix ($\boldsymbol{x}$), in accordance with the methodology outlined in \citet{marinho2025energyindicators}. To ensure uniformity across features, min-max normalisation is applied, constraining each input dimension to [0, 1]. The dataset is partitioned according to the guidelines in \nameref{architecture_NN}, allocating 80\% of the samples to training and 20\% to testing. Model training and evaluation occur within the disjoint PINN framework.

Elaborating on this foundational pipeline, the specific data structuring for each analytical objective is detailed. Case P1 explores predictive analytics with a focus on overall predictive capability under realistic measurement variability encountered in SHM applications. In this scenario, the complete dataset is used, augmented with 1\% Gaussian noise, yielding a total of 146 impact responses for modelling. The augmented dataset, therefore, contains equal proportions of original and noise-perturbed signals. Here, the term \textit{prediction} refers to model-estimated outputs evaluated on samples not provided as labelled targets during the training phase. As such, all quantities inferred from test samples are considered predictions, as they are estimated using the learned surrogate models rather than direct observation.

Cases R1-R3 assess robustness against data limitations and increasing noise, which commonly arise in practical SHM deployments due to limited sensor coverage, constrained testing campaigns, and variable signal quality. Case R1 explores data availability ranging from 25\% to 100\% using only original signals, thereby isolating the effects of reduced sample size. The percentages denote fractions of the total number of impact tests, which consists of 73 impact samples. After the initial 80/20 split, the test set is fixed across all dataset sample sizes, while subsetting is applied only to the training set.

In Case R2, the complete sample set (100\%) is retained while noise levels are incrementally increased from 0\% to 5\%, thereby assessing the robustness against uncertainties in measurement. Case R3 modifies both data availability and noise levels, mirroring the approach of R1, which subsamples the training dataset while keeping the test set fixed. Figure \ref{fig:example_noise} shows an example of a time series signal with the maximum 5\% noise level introduced in Cases R2 and R3, confirming the challenging measurement conditions represented in these cases. Performance metrics across all robustness cases are computed as the average error over five folds, ensuring that the reported trends reflect consistent behaviour rather than relying on a single data split.

Finally, Case G1 investigates generalisation through out-of-distribution testing samples. This process involves defining training events within a specified nominal energy range while keeping impacts outside this range for testing only. Such an approach is crucial for SHM systems operating in environments where new, yet physically related, scenarios may arise without immediate access to additional training data.
%
\section{Results and Discussion} \label{sec:results}
\noindent
The results are presented in the same structure as the data processing workflow defined in the previous section. First, feature extraction and selection are applied to the impact test data to define physics-based energy indicators. Next, results from the predictive analytics case (P1) quantify the accuracy of the estimated mass, velocity, and impact energy at test samples. The analysis then examines robustness through Cases R1–R3, followed by an assessment of generalisation in Case G1 using out-of-distribution scenarios.
\begin{table*}
    \caption{Feature ranking and selection for defining the physics-based energy indicators.}
    \centering
    \small
\begin{tabular}{p{1.75cm}p{1cm}p{1cm}p{0.7cm}p{0.75cm}p{0.75cm}cc}
\toprule
\textbf{Domain} & \textbf{Group} & \textbf{ID} & $\mathbf{w^m}$ & $\mathbf{r^m}$ & $\mathbf{s^m}$ &  \textbf{evaluation} \\
\midrule
\textbf{Time} & T1 & RMS & 0.58 & 0.97 & 0.57 &  $\ast\ast$ / $\bullet\bullet$ \\
  & T2 & TE  & 0.59 & 0.95 & 0.56 & $\ast\ast$ / $\bullet\bullet$ \\
  & T3 & PA  & 0.61 & 0.99 & 0.60 &  $\ast\ast$ / $\bullet\bullet$ \\
  & T4 & EPR & 0.61 & 0.95 & 0.58 &  $\ast\ast$ / $\bullet\bullet$ \\
  & T5 & RA  & 0.34 & 0.80 & 0.28 &  $\ast\ast$ / $\diamond$ \\
\textbf{Frequency}
  & F1 & PCR  & 0.34 & 0.81 & 0.28 &  $\ast\ast$ / $\bullet\bullet$ \\
  & F2 & WPF  & 0.55 & 0.81 & 0.45 &  $\ast\ast$ / $\bullet\bullet$ \\
  & F3 & PF   & 0.52 & 1.00 & 0.52 &  $\ast\ast$ / $\bullet\bullet$ \\
\mbox{\textbf{Time-}}
& W1 & AME & 0.59 & 0.99 & 0.59 &  $\ast\ast$ / $\bullet\bullet$ \\
  \mbox{\textbf{Frequency}}& W2 & AM  & 0.61 & 0.99 & 0.60 & $\ast\ast$ / $\bullet\bullet$ \\
  & W3 & DM  & 0.46 & 0.56 & 0.26 &  $\ast\ast$ / $\diamond$ \\
\bottomrule
\end{tabular}
\vspace{0.5em}
\begin{tablenotes}
\small
\item \textbf{Legend}
\vspace{0.5em}
\item \textbf{Scores}: $\mathbf{w^{(m)}}$ importance score;  $\mathbf{r^{(m)}}$ robustness score;  $\mathbf{s^{(m)}}$ selection score. 
\vspace{0.5em}
\item \textbf{Evaluation}:  $\diamond$ not stable under noisy conditions; $\ast\ast$ independent; $\bullet\bullet$ relevant and stable feature.
\end{tablenotes}
\label{feature_ranking_element}
\end{table*}

\subsection{Physics-based energy indicators} \label{sec:energy_indicators}
\noindent 
This subsection addresses the feature extraction and selection from the impact responses in the case study described in \nameref{sec:dataset}. The selection follows the framework reported in \citet{marinho2025energyindicators} on energy indicators for impact energy estimation in composites. Within this framework, the multi-domain candidate set is assessed using three criteria that characterise reliable indicators: sensitivity to variations in impact energy, robustness under noisy measurements and independence from other descriptors. These criteria ensure that the selected features reflect measurable responses linked to impact mechanics, embedding observational bias within the Phy-ID framework (refer to Part I in \Cref{fig:general_framework_Phy3ID}).

\Cref{feature_ranking_element} summarises the resulting scores and determines the final set of indicators used in this study. Based on these scores, the selected set includes the signal features: Root Mean Square (RMS)~\cite{jang2015impact}, Transmitted Energy (TE)~\cite{tabian2019convolutional}, Peak Amplitude (PA)~\cite{ghajari2013identification}, Energy Peak Ratio (EPR)~\cite{guel2020data}, Peak Centroid Ratio (PCR)~\cite{muir2021damage}, Weighted Peak Frequency (WPF)~\cite{ali2019microscopic}, Peak Frequency (PF)~\cite{kostopoulos2003identification}, Approximation Max Energy (AME)~\cite{oskouei2009wavelet,fotouhi2015investigation} and Approximation Max (AM)~\cite{oskouei2009wavelet,fotouhi2015investigation}. These indicators constitute the inputs, i.e., the feature matrix, for the predictive models assessed in the subsequent analyses. While this section focuses on the final results and the chosen feature set, further methodological details are available in \citet{marinho2025energyindicators}.
\subsection{Predictive analytics (Case P1)}
Case P1 evaluates the predictive capabilities of surrogate models in the context of real-world measurement variability, focusing on inferring impact velocity, impactor mass, and impact energy. Model performance is quantified using the mean absolute percentage error (MAPE) \citep{de_Myttenaere_2016} and further assessed qualitatively by examining the concentration of predictions within the $\pm$10\% tolerance bands, adopted here as a practical engineering threshold for SHM capability in line with recent aerospace studies reporting prediction errors within this range as acceptable performance \citep{Nicassio_2025}.

\begin{figure}
    \centering
    \begin{subfigure}[t]{0.325\textwidth}
        \includegraphics[width=\linewidth]{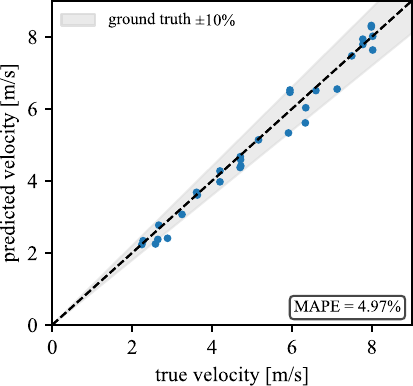}
        \caption{Surrogate prediction of impact velocity.}
        \label{fig:velocity_prediction}
    \end{subfigure}
    \hfill
    \begin{subfigure}[t]{0.325\textwidth}
        \includegraphics[width=\linewidth]{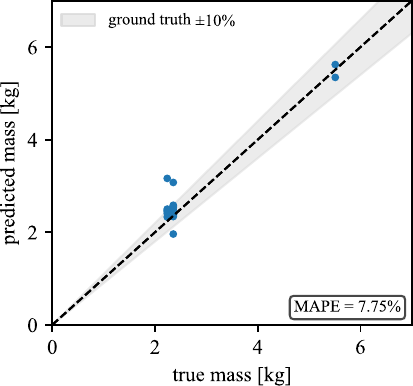}
        \caption{Surrogate prediction of impactor mass.}
        \label{fig:mass_prediction}
    \end{subfigure}
    \hfill
    \begin{subfigure}[t]{0.325\textwidth}
        \includegraphics[width=\linewidth]{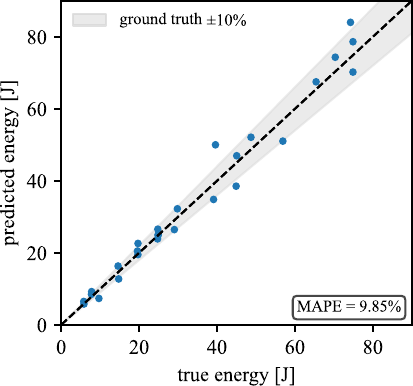}
        \caption{Impact energy inferred via kinetic energy consistency using predicted mass and velocity.}
        \label{fig:energy_prediction}
    \end{subfigure}
    \caption{Predictive analytics results for Case P1.\\
    \footnotesize MAPE: Mean Absolute Percentage Error}
    \label{fig:predictions}
\end{figure}

As illustrated in \Cref{fig:velocity_prediction} and \Cref{fig:mass_prediction}, parity plots depict the surrogate predictions for impact velocity and impactor mass, respectively. The displacement surrogate, which infers impact velocity, achieves a MAPE of 4.97\%, while the mass surrogate reports a MAPE of 7.75\%. Most predictions lie within the shaded $\pm$10\% tolerance band around the ground truth for test samples. For the mass surrogate, however, a few predictions at the 2.238~kg level lie outside the $\pm$10\% band. This deviation likely reflects the limited number of samples available at this mass level and the small difference between the 2.238~kg and 2.356~kg masses. Under certain impact conditions, these two masses generate similar structural responses, increasing the sensitivity of the inverse mass estimation to measurement noise and optimisation trade-offs. Overall, the deviations remain limited and do not alter the general trend observed in the parity plot.

\Cref{fig:energy_prediction} presents the results for impact energy. Impact energy is inferred from the kinetic energy relation (\Cref{kinetic_energy}) using the predicted impactor mass and impact velocity. The resulting estimates yield a MAPE of 9.85\%. Importantly, 88.9\% of the test samples (23 out of 27) fall within the $\pm$10\% tolerance band, confirming that the accuracy of the individual surrogates translates into reliable energy estimation when kinetic consistency is enforced.

\begin{figure}
\centering
\includegraphics[width=0.325\textwidth]{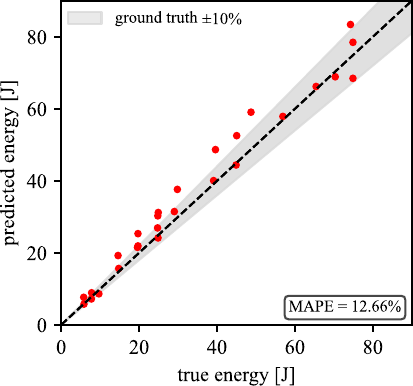}
\caption{Impact energy inferred after removing architectural priors, surrogate components, and hybrid loss terms.}
\label{fig:parity_energy_CR}
\end{figure}

To isolate the contribution of the physics-motivated biases, \Cref{fig:parity_energy_CR} reports the impact energy predictions obtained after removing the physics-informed components while retaining the same FCNN architecture and identical test samples used in the Phy-ID evaluation. In this configuration, the observational bias (OB.1) remains unchanged, whereas inductive biases (IB.1 and IB.3) and learning bias (LB.1) are removed by replacing the physics-informed formulation with a single FCNN trained using a standard mean-squared error loss to directly predict impact energy. Under these conditions, the MAPE increases to 12.66\%. Despite the relatively small difference in average error compared to the physics-informed design (9.85\%), the accuracy levels drop significantly; only 51.8\% of predictions (14 out of 27) remain within the $\pm$10\% threshold. This disparity highlights the increased noise in the predictions when prior knowledge constraints are absent.

Overall, these results indicate that including physics-motivated constraints narrows the feasible solution space and supports more stable energy estimates, which is associated with improved accuracy. Specifically, including prior knowledge increases the proportion of predictions within a defined target tolerance band, demonstrating a positive effect on predictive accuracy under the same architecture and test conditions.

\subsection{Robustness studies (Cases R1-R3)}
\noindent
The robustness evaluation of the proposed framework is detailed in Cases R1–R3 with a particular focus on impact energy prediction. These analyses investigate the performance of the model under conditions of reduced data availability and degraded measurement quality. The mean absolute percentage error (MAPE) is utilised to quantify robustness, with the average computed across five folds to ensure reliable performance.

\begin{figure*}
    \centering
    \includegraphics[width=\linewidth]{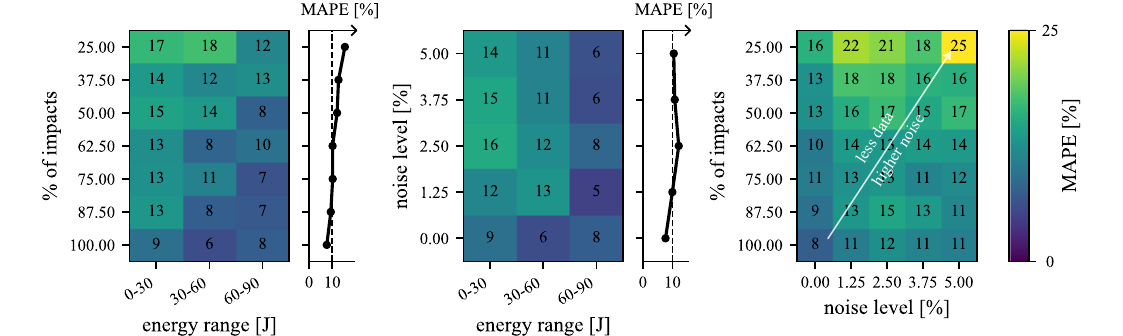}
    \caption{Impact energy robustness analysis. Measured error across energy bins for (left) varying data availability at zero noise (Case R1), (centre) increasing noise levels at full data availability (Case R2), and (right) combined reductions in data availability and increases in noise (Case R3).\\
    \footnotesize MAPE: Mean Absolute Percentage Error}
    \label{fig:sensitivity_analysis}
\end{figure*}

\Cref{fig:sensitivity_analysis} summarises the results, presenting heatmap plots that illustrate MAPE values across energy bins for different scenarios, and measured average error for all energy levels in combined scenarios. Each plot features numerical annotations that specify corresponding error metrics. In Cases R1 and R2, adjacent plots show error metrics averaged over all energy levels, enabling a comprehensive assessment of overall performance behaviour.

In Case R1, illustrated in the left panel of Figure \ref{fig:sensitivity_analysis}, predictive performance exhibits limited sensitivity to reductions in data availability. MAPE values, averaged across all energy levels, remain closely aligned at approximately 10\% as data availability decreases from 100\% to around 62.5\%. A slight increase in average error is observed as training sample size decreases, particularly between 50\% and 25\% data availability. Notably, even at 25\% data availability, representing only 15 impact samples for training, the average MAPE sustains at approximately 15\%. From a SHM perspective, such accuracy remains significant and practical for warning systems, depending on the context and risk tolerance~\citep{Moradi_2023,Ofir_2025,Broer_2021}.

When examining individual energy ranges, the analysis reveals more pronounced variations in MAPE values at lower impact energies, which are influenced by the formulation of the error metric. As MAPE accounts for percentage error relative to true values, small absolute residuals in low-energy ranges can inflate percentage errors considerably~\citep{Tofallis_2015,Kim_2016}. Thus, the relative increase in MAPE at limited sample sizes reflects both reduced training information and greater metric sensitivity in lower-magnitude regions.

Case R2, depicted in the central panel of Figure \ref{fig:sensitivity_analysis}, isolates sensitivity to measurement noise, using the complete training samples while varying noise levels from 0\% to 5\%. Under these conditions, the average MAPE remains close to 10\%, with no clear systematic dependence on impact energy. During input space design (see previous section \nameref{sec:energy_indicators}), the selected descriptors were evaluated for both energy sensitivity and robustness under controlled noise conditions. Because robustness to noise was one of the criteria guiding the construction of the input space, the limited variation in error under moderate noise levels is consistent with the feature design strategy adopted in this work.

 Comparison of cases R1 and R2 indicates that reductions in data availability have a greater impact on model performance than higher noise levels. This trend aligns with established insights regarding learning-based models, where less training information has a more pronounced effect on robustness than measurement noise \citep{Miele_2023,Zheng_2025,Wang_2024}. 

Case R3, illustrated in the right panel of Figure \ref{fig:sensitivity_analysis}, integrates the combined effect of reduced data availability and increased noise levels. Performance remains stable, with average MAPE values around 15\% across most combinations. It is only under the most extreme conditions, 25\% data availability paired with 5\% noise, that MAPE escalates to 25.4\%. This observation indicates a gradual, expected performance decline rather than a sudden failure of the model's capabilities. From a general perspective, the robustness analyses show that the proposed Phy-ID framework preserves predictive performance under varying levels of data availability and measurement noise. 

\subsection{Generalisation assessment (Case G1)}
\noindent
The generalisation assessment examines the ability of the proposed Phy-ID framework to extrapolate impact energy predictions beyond the training domain. Models are trained on restricted nominal energy intervals and assessed exclusively on out-of-distribution (OOD) samples that lie outside these intervals.

\begin{figure*}
    \centering
    \begin{subfigure}[t]{0.475\textwidth}
        \includegraphics[width=\linewidth]{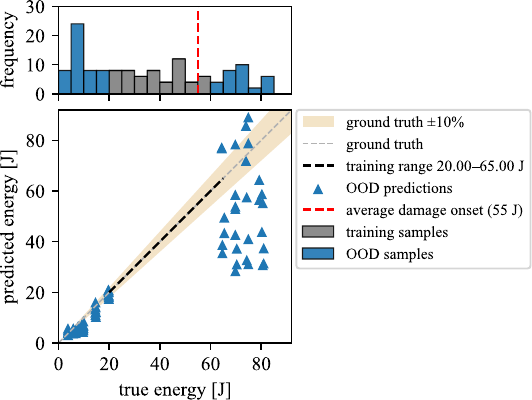}
        \caption{Energy estimation for a training range of 20–65~J; including 25\% of damaged samples.}
        \label{fig:generalisation_ood_training_20_64}
    \end{subfigure}
    \qquad
    \begin{subfigure}[t]{0.475\textwidth}
        \includegraphics[width=\linewidth]{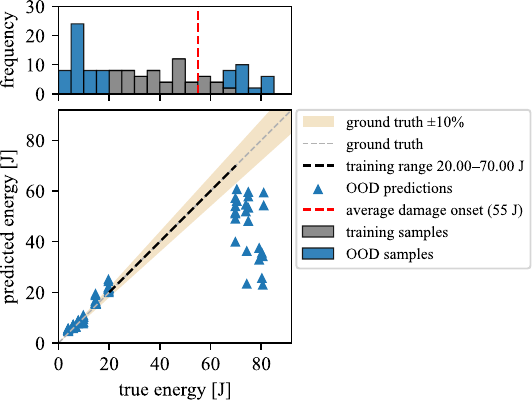}
        \caption{Energy estimation for a training range of 20–70~J; including 40\% of damaged samples.}
        \label{fig:generalisation_ood_training_20_70}
    \end{subfigure}
     \vspace{1em}
    \begin{subfigure}[t]{0.475\textwidth}
        \includegraphics[width=\linewidth]{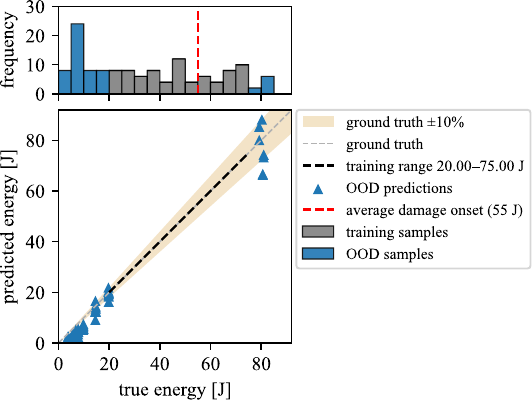}
        \caption{Energy estimation for a training range of 20–75~J; including 60\% of damaged samples.}
        \label{fig:generalisation_ood_training_20_75}
    \end{subfigure}
    \qquad
    \begin{subfigure}[t]{0.475\textwidth}
        \includegraphics[width=\linewidth]{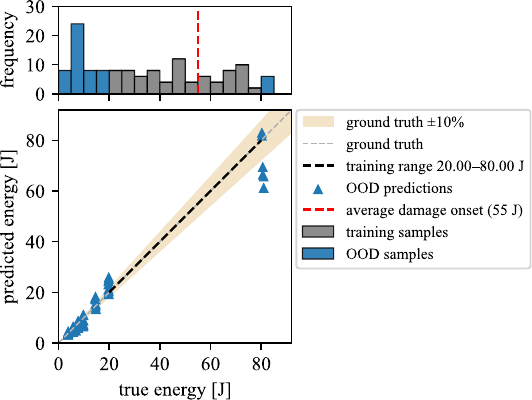}
        \caption{Energy estimation for a training range of 20–80~J; including 80\% of damaged samples.}
        \label{fig:generalisation_ood_training_20_80}
    \end{subfigure}
    \hfill

    \caption{Generalisation behaviour for varying training intervals.  
    Predictions are evaluated on Out-Of-Distribution (OOD) samples outside each training range, illustrating extrapolation in impact energy estimation.}
    \label{fig:generalisation_ood}
\end{figure*}

\Cref{fig:generalisation_ood} summarises the extrapolation behaviour for four progressively extended training intervals. Each subfigure shows parity plots of predicted versus true impact energy for OOD samples. Predictions at low impact energies (0–20~J) consistently demonstrate accuracy and stability, independent of the selected training range. This consistency reflects the inherent stability of the underlying impact response in the pristine regime, indicating that the framework generalises reliably within this linear, damage-free domain.

In contrast, extrapolations at higher impact energies exhibit a marked dependence on the proportion of damaged data included in the training set. In  \Cref{fig:generalisation_ood_training_20_64}, where the training range spans 20–65~J with approximately 25\% of samples damaged, predictions at higher energies are widely scattered and poorly aligned with the parity line. This may indicate that insufficient information exists to capture the altered response associated with damage progression.

With an extended training range of 20–70~J, corresponding to approximately 40\% of damaged samples (\Cref{fig:generalisation_ood_training_20_70}), predictions at higher energies become more clustered and trend towards the ground-truth line. Despite this improvement, noticeable dispersion persists, and many predictions remain outside the $\pm10\%$ bounds, suggesting incomplete learning of the damaged-state response.

A significant qualitative shift occurs when the training set exceeds 50\% of damaged samples. In \Cref{fig:generalisation_ood_training_20_75}, where the training interval is 20–75~J and contains approximately 60\% of damaged samples, OOD predictions at higher energies align more closely with the parity line. This behaviour is maintained with further inclusion of damaged data, as illustrated in \Cref{fig:generalisation_ood_training_20_80}. The predictions qualitatively improve extrapolation performance compared with narrower training intervals, although some points remain outside the $\pm$10\% band. This indicates that increasing the proportion of damaged samples improves trend consistency through better data coverage, but does not eliminate residual scatter. The remaining dispersion likely reflects either the increased nonlinearity of the damaged regime, measurement variability, or the limited physical description of damage progression in the present model.

This trend directly relates to the change in structural response associated with damage onset. As identified in \nameref{sec:dataset} through ultrasonic inspections, damage onset occurs at around 55~J. Below this threshold, the impact response remains predominantly linear, whereas above it, the presence of damage introduces a more nonlinear relationship between impact energy and the measured response. The Phy-ID framework incorporates observational, inductive, and learning biases to encode physically meaningful relationships for impact energy inference. However, it does not explicitly integrate models of damage mechanics or failure progression. Consequently, the increased nonlinearity associated with damage must be learned solely from the data.

\begin{figure}
\centering
\includegraphics[width=0.475\textwidth]{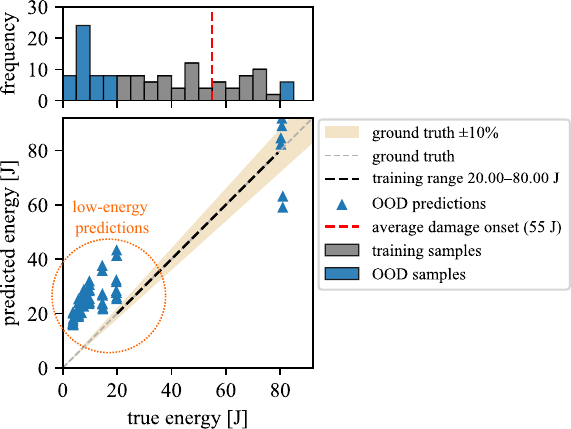}
\caption{Generalisation behaviour without physics-motivated architectural priors, surrogate components, and hybrid loss terms. Predictions on out-of-distribution samples illustrate extrapolation in impact energy estimation.}
\label{fig:generalisation_CR}
\end{figure}

To further assess extrapolation behaviour, \Cref{fig:generalisation_CR} shows the OOD energy predictions from a model trained over the same 20–80~J interval (and 80\% of the damaged samples) but without the physics-informed biases. Comparison with \Cref{fig:generalisation_ood_training_20_80} reveals that in the low-energy range (0–20~J), the Phy-ID formulation maintains predictions well-aligned with the ground truth, whereas the unconstrained model exhibits greater spread and visible deviations from the parity line. At higher impact energies, both formulations show comparable behaviour once a sufficient proportion of damaged samples is included during training. These observations indicate that incorporating prior physical structure improves prediction consistency in the linear regime, while performance in the damaged regime reflects the level of data representation in the training set under the present formulation.

Taken together, these results indicate that the framework achieves limited extrapolation capability for impact energy estimation, provided that a minimum amount of damaged data is available. Generalisation across damage states, therefore, requires access to representative damaged responses during training. In this context, further improvements may be achieved by extending the Phy-ID framework to incorporate physical constraints on damage mechanisms. Such extensions would maybe reduce reliance on damaged samples alone and enhance generalisation across structural states.

\subsection{Discussion}
\noindent
The preceding results, together with characteristics reported in the literature, enable a qualitative comparison between data-driven, physics-based, probabilistic, and physics-informed approaches for impact energy estimation under realistic SHM conditions characterised by low-quality measurements and incomplete physical knowledge. The comparison focuses on four criteria that directly affect deployability in aerospace structures: accuracy, robustness, generalisation, and interpretability. The interpretation of these findings draws on established characteristics of each modelling paradigm reported in the literature and contrasts them with the behaviour observed for the physics-informed formulation in this study. The present comparison, therefore, remains qualitative and context-specific.
\begin{table*}
\centering
\caption{Qualitative comparison of impact identification strategies under limited and noisy data with incomplete physics.}
\label{tab:method_comparison}
\small
\begin{tabular}{lcccc}
\toprule
\textbf{Criterion} & \textbf{Data-driven} & \textbf{Physics-based} & \textbf{Probabilistic} & \textbf{Physics-informed (Phy-ID)} \\
\midrule
Accuracy & 
Low--Medium~\citep{Malekloo_2021,Azimi_2020,Jia_2023} & 
Low--Medium~\citep{Kralovec_2020,Fan_2021} & 
Medium~\citep{Blakseth_2022,Arias_Chao_2022} & 
High \\
Robustness & 
Low~\citep{Azimi_2020,Nicassio_2025} & 
Low--Medium~\citep{Fan_2021, Hassani_2021} & 
Medium~\citep{Blakseth_2022,Arias_Chao_2022} & 
High \\
Generalisation & 
Low~\citep{Azimi_2020,Nicassio_2025} & 
Low--Medium~\citep{Fan_2021, Malekloo_2021} & 
Medium~\citep{Blakseth_2022,Arias_Chao_2022} & 
Medium \\
Interpretability & 
Low~\citep{Azimi_2020,Nicassio_2025} & 
High~\citep{Fan_2021, Malekloo_2021} & 
Low--Medium~\citep{Blakseth_2022,Arias_Chao_2022} & 
High \\
\bottomrule
\end{tabular}
\end{table*}

Table~\ref{tab:method_comparison} summarises this qualitative assessment. It reflects the practical SHM setting addressed here and should not be interpreted as a general ranking across all applications. The performance levels assigned to the data-driven, physics-based, and probabilistic approaches are grounded in the literature cited in the table, whereas the levels attributed to the physics-informed formulation follow directly from the results presented in this study. 

Purely data-driven approaches learn mappings directly from measured data without embedding governing relations. The SHM literature consistently links these methods to strong dependence on large and representative datasets, sensitivity to measurement noise, and weak extrapolation beyond the training domain~\citep{Azimi_2020,Malekloo_2021}. Although good accuracy can be achieved under controlled conditions~\citep{Nicassio_2025}, reliability decreases when operating conditions vary or data coverage is restricted. Without physical constraints, predictions may remain statistically consistent while lacking mechanical plausibility, which limits interpretability and generalisation~\citep{Azimi_2020,Jia_2023}. These characteristics correspond to the low-to-medium accuracy, low robustness, low generalisation and low interpretability reflected in Table~\ref{tab:method_comparison}.

In contrast, physics-based approaches formulate the problem through governing equations and constitutive relations. This structure provides high interpretability and supports extrapolation within the assumed physics. However, performance depends directly on model fidelity. Reviews highlight sensitivity to environmental variability, modelling simplifications, and incomplete representation of damage mechanisms~\citep{Kralovec_2020,Hassani_2021}. In addition, high-fidelity formulations can become computationally demanding, limiting scalability and real-time use~\citep{Malekloo_2021,Fan_2021}. Consequently, accuracy and robustness are often characterised as low to medium, and generalisation remains restricted to explicitly modelled phenomena.

Probabilistic approaches combine prior assumptions with statistical inference to mitigate these limitations. The literature describes them as intermediate solutions whose effectiveness depends on how explicitly physical relations are embedded~\citep{Wang_2025}. When prior information enters indirectly through regularisation rather than enforced constraints, interpretability decreases and inconsistencies may arise~\citep{Ni_2020}. Performance remains conditioned by the quality of prior assumptions and data representativeness, particularly under noisy or limited data conditions~\citep{Hughes_2021}. Increased computational demand may further limit scalability. These factors place probabilistic methodologies at intermediate levels of accuracy, robustness, and interpretability, with moderate generalisation shaped by system-specific assumptions.

The proposed Phy-ID framework integrates domain knowledge at three levels: input definition, architectural formulation, and loss design. Physics-informed learning literature associates explicit enforcement of governing relations with improved tolerance to limited and noisy data and with stronger interpretability~\citep{Cuomo_2022, Wu_2024,Xu_2023}. By restricting admissible solutions to those consistent with known physical descriptions, such formulations moderate solution variability. 

The behaviour observed for the proposed Phy-ID framework aligns with the conceptual expectations reported in the literature and demonstrates their practical relevance. In Case~P1, the physics-informed configuration achieves a higher concentration of predictions within the tolerance band than the unconstrained variant. In Cases~R1–R3, performance degrades gradually under reduced data and increased noise, reflecting enhanced robustness. In Case~G1, extrapolation remains consistent within the linear regime, while behaviour at higher energies depends on the availability of representative samples, indicating moderate generalisation. In terms of interpretability, the disjoint inference of impactor mass and velocity exposes physically meaningful intermediate quantities instead of relying solely on direct energy regression. Together, these characteristics support the high accuracy, high robustness, high interpretability, and moderate generalisation levels attributed to the physics-informed approach in Table~\ref{tab:method_comparison}.


\section{Concluding remarks} \label{sec:conclusion}
\noindent 
The Phy-ID framework combines observational, inductive, and learning biases, ensuring a balanced approach to impact identification that prioritises physically admissible predictions, data efficiency, and interpretability. This integrated method provides a cohesive solution for monitoring impact events in composite structures, overcoming the limitations of reference methods under realistic conditions.

The results show that the framework infers impact velocity and mass in a physically coherent manner, enabling impact energy estimation through kinetic consistency. The controlled comparison further indicates a positive effect of the imposed physical constraints, reflected in a higher proportion of predictions within the tolerance band under identical conditions. Across the evaluated scenarios, performance remains robust to moderate data reduction and measurement noise. Extrapolation beyond the training range remains feasible in the linear regime. In the damaged regime, however, prediction quality depends on adequate representation of damaged responses in the training set under the present implementation. 

The conclusions are drawn from a case study designed to evaluate the framework under controlled conditions that reproduce key challenges encountered in practical SHM monitoring scenarios. The test campaign involves an element-level composite structure subjected to multiple impact scenarios under challenging measurement conditions. Within this setting, the framework infers physically interpretable quantities, including impact velocity and mass, with quantified error levels. The explicit separation of underlying physical parameters within the formulation provides a structured basis that may allow adaptation to different monitoring objectives and extension to additional impact attributes through the inclusion of relevant physical descriptors and governing relations.

Although the framework demonstrates favourable performance, some aspects require further consideration. First, performance depends on the representativeness of the selected feature set. This assumes that the measured data capture the relevant physical information. As such, structural configurations outside the training envelope may therefore reduce accuracy. Second, the absence of ground-truth measurements for the inferred physical quantities may affect direct validation of the parameter estimates in experimental campaigns where such measurements are impractical. In addition, the lack of explicit constraints related to failure mechanisms limits performance at higher energy levels, where structural behaviour changes due to damage onset and progression. Lastly, measurement uncertainty is not explicitly represented; consequently, confidence bounds on the inferred quantities are unavailable, which may limit applicability in safety-critical contexts.

Future work should address these limitations along three main directions. Transfer learning strategies can enhance scalability and data efficiency. As the framework relies on domain-invariant physical relationships, knowledge transfer across varying structures, geometries, and loading regimes becomes feasible, minimising the necessity for extensive experimental datasets and reducing computational costs. In this context, the framework can further benefit from multi-fidelity formulations, where information from numerical simulations complements experimental observations, improving parameter inference when testing data are scarce, costly to obtain, or lack direct measurements of impact parameters. 

Beyond transfer learning, the framework can be expanded to target additional physical quantities and constraints, such as failure criteria, health indicators, damage initiation thresholds, and contact-related parameters, which directly influence stress distribution and local damage development. Another promising direction involves incorporating uncertainty-aware formulations, such as probabilistic surrogate modelling or Bayesian parameter inference, which explicitly represent measurement uncertainty and confidence in predicted quantities, thereby supporting informed maintenance decisions.

Overall, the proposed Phy-ID framework provides a robust, interpretable, and versatile foundation for impact identification in composite structures. By integrating multiple forms of bias, it delivers stable performance under realistic monitoring conditions and facilitates physically meaningful inference that supports engineering analysis and operational decision-making. The identified limitations and extension pathways highlight how the framework can evolve toward scalable impact monitoring solutions within SHM systems.


\begin{acks}
This work is part of the PrimaVera Project, which is partly financed by the Dutch Research Council (NWO) under grant agreement NWA.1160.18.238.
\end{acks}

\bibliographystyle{unsrtnat} 
\bibliography{bibliography/references}

@article{LIU2023107873,
    author = {Yaru Liu and Lei Wang},
    title = {Quantification, localization, and reconstruction of impact force on interval composite structures},
    journal = {International Journal of Mechanical Sciences},
    year = {2023},
    OPTvolume = {239},
    OPTpages = {107873},
    OPTissn = {0020-7403},
    doi = {10.1016/j.ijmecsci.2022.107873}
}

@article{Correas2021Analytical,
    author = {A. C. Correas and A. Crespo and H. Ghasemnejad and G. Roshan},
    title = {Analytical Solutions to Predict Impact Behaviour of Stringer Stiffened Composite Aircraft Panels},
    journal = {Applied Composite Materials},
    year = {2021},
    OPTvolume = {28},
    OPTpages = {1237 - 1254},
    doi = {10.1007/s10443-021-09909-8}
}

@article{tabian2019convolutional,
    author = {Tabian, Iuliana and Fu, Hailing and Sharif Khodaei, Zahra},
    title = {A Convolutional Neural Network for Impact Detection and Characterization of Complex Composite Structures},
    journal = {Sensors},
    year = {2019},
    OPTvolume = {19},
    OPTnumber = {22},
    OPTpubmedid = {31726762},
    OPTissn = {1424-8220},
    doi = {10.3390/s19224933}
}

@article{sharif2013smart,
    author = {Sharif-Khodaei, Z and Ghajari, M and MH Aliabadi, Ferri and Apicella, A},
    title = {{SMART} platform for Structural Health Monitoring of sensorised stiffened composite panels},
    journal = {Key Engineering Materials},
    year = {2013},
    OPTvolume = {525},
    OPTpages = {581--584},
    OPTpublisher = {Trans Tech Publ},
    doi = {10.4028/www.scientific.net/KEM.525-526.581}
}

@article{yan2017impact,
    author = {Yan, Gang and Sun, Hao and Büyüköztürk, Oral},
    title = {Impact load identification for composite structures using Bayesian regularization and unscented Kalman filter},
    journal = {Structural Control and Health Monitoring},
    year = {2017},
    OPTvolume = {24},
    OPTnumber = {5},
    OPTpages = {e1910},
    doi = {10.1002/stc.1910}
}

@article{ZHANG2020111882,
    author = {Xiaoyu Zhang and Fei Xu and Yuyan Zang and Wei Feng},
    title = {Experimental and numerical investigation on damage behavior of honeycomb sandwich panel subjected to low-velocity impact},
    journal = {Composite Structures},
    year = {2020},
    OPTvolume = {236},
    OPTpages = {111882},
    ISSN = {0263-8223},
    doi = {10.1016/j.compstruct.2020.111882}
}

@article{karniadakis2021physics,
    author = {Karniadakis, George Em and Kevrekidis, Ioannis G and Lu, Lu and Perdikaris, Paris and Wang, Sifan and Yang, Liu},
    title = {Physics-informed machine learning},
    journal = {Nature Reviews Physics},
    year = {2021},
    OPTvolume = {3},
    OPTnumber = {6},
    OPTpages = {422--440},
    OPTpublisher = {Nature Publishing Group UK London},
    doi = {10.1038/s42254-021-00314-5}
}

@article{Cross_Rogers_Pitchforth_Gibson_Zhang_Jones_2024,
    author = {Cross, Elizabeth J. and Rogers, Timothy J. and Pitchforth, Daniel J. and Gibson, Samuel J. and Zhang, Sikai and Jones, Matthew R.},
    title = {A spectrum of physics-informed Gaussian processes for regression in engineering},
    journal = {Data-Centric Engineering},
    year = {2024},
    OPTvolume = {5},
    OPTpages = {e8},
    doi = {10.1017/dce.2024.2}
}

@article{AMININIAKI2021113959, 
    title={Physics-informed neural network for modelling the thermochemical curing process of composite-tool systems during manufacture}, 
    OPTvolume={384}, 
    OPTissn={0045-7825}, 
    DOI={10.1016/j.cma.2021.113959}, 
    journal={Computer Methods in Applied Mechanics and Engineering}, 
    publisher={Elsevier}, 
    author={Amini Niaki, Sina and Haghighat, Ehsan and Campbell, Trevor and Poursartip, Anoush and Vaziri, Reza}, 
    year={2021}, 
    OPTpages={113959} 
}

@article{olsson2000mass,
    author = {Olsson, R.},
    title = {Mass criterion for wave controlled impact response of composite plates},
    journal = {Composites Part A: Applied Science and Manufacturing},
    year = {2000},
    OPTvolume = {31},
    OPTnumber = {8},
    OPTpages = {879–887},
    OPTissn = {1359-835X},
    OPTpublisher = {Elsevier},
    doi = {10.1016/s1359-835x(00)00020-8}
}

@book{ASTMD7136,
    author = {ASTM},
    title = {{D7136/D7136M – 15 Standard Test Method for Measuring the Damage Resistance of a Fiber-Reinforced Polymer Matrix Composite to a Drop-Weight Impact Event }},
    year = {2015},
    publisher = {American Society for Testing and Materials},
    OPTaddress = {West Conshohocken, PA}
}

@article{yue2021damage, 
    title={Damage detection in large composite stiffened panels based on a novel SHM building block philosophy}, 
    OPTvolume={30}, 
    OPTissn={1361-665X}, 
    DOI={10.1088/1361-665x/abe4b4}, 
    OPTnumber={4}, 
    journal={Smart Materials and Structures}, 
    publisher={{IOP} Publishing}, 
    author={Yue, Nan and Khodaei, Zahra Sharif and Aliabadi, M H}, 
    year={2021},
    OPTpages={045004}
}

@book{faa2009ac,
    author = {U.S. Department of Transportation Federal Aviation Administration ({FAA})},
    title = {Advisory Circular ({AC}) 20-107b Composite Aircraft Structure},
    year = {2010},
    publisher = {American Society for Testing and Materials},
    OPTaddress = {West Conshohocken, PA}
}

@article{marinho2024impact, 
    title={Impact Identification Method for Structural Health Monitoring of Stiffened Composite Panels using Passive Sensing Systems}, 
    OPTvolume={29}, 
    OPTissn={1435-4934}, 
    DOI={10.58286/29655}, 
    OPTnumber={7},
    journal={e-Journal of Nondestructive Testing}, publisher={{NDT}.net {GmbH} \& Co. {KG}}, 
    author={Marinho, Nat\'{a}lia Ribeiro and Loendersloot, Richard and Grooteman, Frank and Wiegman, Jan Willem and Tinga, Tiedo}, 
    year={2024}
}

@article{marinho2025evaluating,
    author = {Marinho, Natália Ribeiro and Loendersloot, Richard and Wiegman, Jan Willem and Grooteman, Frank and Tinga, Tiedo},
    title = {Evaluating sensor performance for impact identification in composites: a comprehensive comparison of {FBGs} with {PZTs}},
    journal = {Structural Health Monitoring},
    year = {2025},
    OPTissn = {1741-3168},
    OPTpublisher = {SAGE Publications},
    doi = {10.1177/14759217241304644}
}

@article{khalid2024advancements,
    author = {Khalid, Salman and Yazdani, Muhammad Haris and Azad, Muhammad Muzammil and Elahi, Muhammad Umar and Raouf, Izaz and Kim, Heung Soo},
    title = {Advancements in Physics-Informed Neural Networks for Laminated Composites: A Comprehensive Review},
    journal = {Mathematics},
    year = {2025},
    OPTvolume = {13},
    OPTnumber = {1},
    OPTarticle-number = {17},
    OPTissn = {2227-7390},
    doi = {10.3390/math13010017}
}

@misc{marinho2025energyindicators,
      title={{Defining Energy Indicators for Impact Identification on Aerospace Composites: A Physics-Informed Machine Learning Perspective}}, 
      author={Nat\'{a}lia Ribeiro Marinho and Richard Loendersloot and Frank Grooteman and Jan Willem Wiegman and Uraz Odyurt and Tiedo Tinga},
      year={2025},
      eprint={2511.01592},
      archivePrefix={arXiv},
      primaryClass={cs.LG},
      url={https://arxiv.org/abs/2511.01592}, 
}

@article{Andrew_2019, 
    title={Parameters influencing the impact response of fiber-reinforced polymer matrix composite materials: A critical review}, 
    OPTvolume={224}, 
    OPTISSN={0263-8223}, 
    DOI={10.1016/j.compstruct.2019.111007}, 
    journal={Composite Structures}, 
    publisher={Elsevier}, 
    author={Andrew, J. Jefferson and Srinivasan, Sivakumar M. and Arockiarajan, A. and Dhakal, Hom Nath}, 
    year={2019}, 
    OPTpages={111007} 
}

@article{Zhang_2024, 
    title={An impact localization method of composite fan blades based on stress wave features}, 
    OPTvolume={34}, 
    OPTISSN={1361-665X}, 
    DOI={10.1088/1361-665x/ad9e5c}, 
    OPTnumber={1}, 
    journal={Smart Materials and Structures}, 
    publisher={{IOP} Publishing}, 
    author={Zhang, Qingchen and Zhao, Bowen and Liu, Qijian and Liu, Hailong and Huang, Meiao and Qing, Xinlin}, 
    year={2024}, 
    OPTpages={015046} 
}

@article{Aryal_2019, 
    title={Effects of impact energy, velocity, and impactor mass on the damage induced in composite laminates and sandwich panels}, 
    OPTvolume={226}, 
    OPTISSN={0263-8223}, 
    DOI={10.1016/j.compstruct.2019.111284}, 
    journal={Composite Structures}, 
    publisher={Elsevier}, 
    author={Aryal, B. and Morozov, E.V. and Wang, H. and Shankar, K. and Hazell, P.J. and Escobedo-Diaz, J.P.}, 
    year={2019}, 
    OPTpages={111284} 
}

@article{Artero_Guerrero_2015, 
    title={Experimental study of the impactor mass effect on the low velocity impact of carbon/epoxy woven laminates},
    OPTvolume={133}, 
    OPTISSN={0263-8223},
    DOI={10.1016/j.compstruct.2015.08.027}, 
    journal={Composite Structures},
    publisher={Elsevier},
    author={Artero-Guerrero, J.A. and Pernas-Sánchez, J. and López-Puente, J. and Varas, D.},
    year={2015}, 
    OPTpages={774–781} 
}

@article{Mukhopadhyay_2020, 
    title={{Stochastic Oblique Impact on Composite Laminates: A Concise Review and Characterization of the Essence of Hybrid Machine Learning Algorithms}}, 
    OPTvolume={28}, 
    OPTISSN={1886-1784}, 
    DOI={10.1007/s11831-020-09438-w}, 
    OPTnumber={3}, 
    journal={Archives of Computational Methods in Engineering}, 
    publisher={{Springer Science and Business Media LLC}},
    author={Mukhopadhyay, T. and Naskar, S. and Chakraborty, S. and Karsh, P. K. and Choudhury, R. and Dey, S.},
    year={2020}, 
    OPTpages={1731–1760} 
}

@article{Wang_2021, 
    title={{Understanding and Mitigating Gradient Flow Pathologies in Physics-Informed Neural Networks}}, 
    OPTvolume={43}, 
    OPTISSN={1095-7197},
    DOI={10.1137/20m1318043}, 
    OPTnumber={5}, 
    journal={SIAM Journal on Scientific Computing}, 
    publisher={Society for Industrial \& Applied Mathematics (SIAM)}, 
    author={Wang, Sifan and Teng, Yujun and Perdikaris, Paris}, 
    year={2021}, 
    OPTpages={A3055–A3081} 
}

@article{Perez_2023, 
    title={Adaptive weighting of Bayesian physics-informed neural networks for multitask and multiscale forward and inverse problems}, 
    OPTvolume={491}, 
    OPTISSN={0021-9991}, 
    DOI={10.1016/j.jcp.2023.112342}, 
    journal={{Journal of Computational Physics}}, 
    publisher={Elsevier}, 
    author={Perez, Sarah and Maddu, Suryanarayana and Sbalzarini, Ivo F. and Poncet, Philippe}, 
    year={2023}, 
    OPTpages={112342}
}

@article{Xiang_2022, 
    title={Self-adaptive loss balanced Physics-informed neural networks}, 
    OPTvolume={496}, 
    OPTISSN={0925-2312}, 
    DOI={10.1016/j.neucom.2022.05.015}, 
    journal={Neurocomputing}, 
    publisher={Elsevier}, 
    author={Xiang, Zixue and Peng, Wei and Liu, Xu and Yao, Wen}, 
    year={2022},
    OPTpages={11–34} 
}

@article{jang2015impact,
    author = {Byeong-Wook Jang and Yeon-Gwan Lee and Chun-Gon Kim and Chan-Yik Park},
    title = {Impact source localization for composite structures under external dynamic loading condition},
    journal = {Advanced Composite Materials},
    year = {2015},
    OPTvolume = {24},
    OPTnumber = {4},
    OPTpages = {359--374},
    OPTpublisher = {Taylor \& Francis},
    doi = {10.1080/09243046.2014.917239}
}

@article{ghajari2013identification,
    author = {Ghajari, M and Sharif-Khodaei, Z and Aliabadi, M H and Apicella, A},
    title = {Identification of impact force for smart composite stiffened panels},
    journal = {Smart Materials and Structures},
    year = {2013},
    OPTvolume = {22},
    OPTnumber = {8},
    OPTpages = {085014},
    OPTmonth = {jul},
    OPTpublisher = {IOP Publishing},
    doi = {10.1088/0964-1726/22/8/085014}
}

@article{guel2020data,
    author = {Guel, Nicolas and Hamam, Zeina and Godin, Nathalie and Reynaud, Pascal and Caty, Olivier and Bouillon, Florent and Paillassa, Aude},
    title = {Data Merging of {AE} Sensors with Different Frequency Resolution for the Detection and Identification of Damage in Oxide-Based Ceramic Matrix Composites},
    journal = {Materials},
    year = {2020},
    OPTvolume = {13},
    OPTnumber = {20},
    OPTarticle-number = {4691},
    OPTissn = {1996-1944},
    OPTpublisher = {MDPI},
    doi = {10.3390/ma13204691}
}

@article{muir2021damage,
    author = {Muir, C and Swaminathan, B and Almansour, AS and Sevener, K and Smith, C and Presby, M and Kiser, JD and Pollock, TM and Daly, S},
    title = {Damage mechanism identification in composites via machine learning and acoustic emission},
    journal = {{NPJ} Computational Materials},
    year = {2021},
    OPTvolume = {7},
    OPTnumber = {1},
    OPTpages = {95},
    OPTpublisher = {Nature Publishing Group UK London},
    doi = {10.1038/s41524-021-00565-x}
}

@article{ali2019microscopic,
    author = {Hafiz Qasim Ali and Isa {Emami Tabrizi} and Raja Muhammad Awais Khan and Ali Tufani and Mehmet Yildiz},
    title = {Microscopic analysis of failure in woven carbon fabric laminates coupled with digital image correlation and acoustic emission},
    journal = {Composite Structures},
    year = {2019},
    OPTvolume = {230},
    OPTpages = {111515},
    OPTissn = {0263-8223},
    doi = {10.1016/j.compstruct.2019.111515}
}

@article{kostopoulos2003identification,
    author = {V Kostopoulos and T.H Loutas and A Kontsos and G Sotiriadis and Y.Z Pappas},
    title = {On the identification of the failure mechanisms in oxide/oxide composites using acoustic emission},
    journal = {{NDT} \& E International},
    year = {2003},
    OPTvolume = {36},
    OPTnumber = {8},
    OPTpages = {571-580},
    OPTissn = {0963-8695},
    doi = {10.1016/S0963-8695(03)00068-9}
}

@article{oskouei2009wavelet,
    author = {Oskouei, A Refahi and Ahmadi, M and Hajikhani, M},
    title = {Wavelet-based acoustic emission characterization of damage mechanism in composite materials under {Mode I} delamination at different interfaces},
    journal = {Express Polymer Letters},
    year = {2009},
    OPTvolume = {3},
    OPTnumber = {12},
    OPTpages = {804--813},
    doi = {10.3144/expresspolymlett.2009.99}
}

@article{fotouhi2015investigation,
    author = {Mohamad Fotouhi and Milad Saeedifar and Seyedali Sadeghi and Mehdi Ahmadi Najafabadi and Giangiacomo Minak},
    title = {Investigation of the damage mechanisms for {Mode I} delamination growth in foam core sandwich composites using acoustic emission},
    journal = {Structural Health Monitoring},
    year = {2015},
    OPTvolume = {14},
    OPTnumber = {3},
    OPTpages = {265-280},
    doi = {10.1177/1475921714568403}
}

@article{de_Myttenaere_2016, 
    title={{Mean Absolute Percentage Error for regression models}}, 
    OPTvolume={192}, 
    OPTISSN={0925-2312}, 
    DOI={10.1016/j.neucom.2015.12.114}, 
    journal={Neurocomputing}, 
    publisher={{Elsevier BV}}, 
    author = {de Myttenaere, Arnaud and Golden, Boris and Le Grand, Bénédicte and Rossi, Fabrice}, 
    year={2016},
    OPTpages={38–48} 
}

@article{Moradi_2023, 
    title={Intelligent health indicator construction for prognostics of composite structures utilizing a semi-supervised deep neural network and SHM data},
    OPTvolume={117}, 
    OPTISSN={0952-1976},
    DOI={10.1016/j.engappai.2022.105502}, 
    journal={Engineering Applications of Artificial Intelligence},
    publisher={{Elsevier BV}}, 
    author={Moradi, Morteza and Broer, Agnes and Chiachío, Juan and Benedictus, Rinze and Loutas, Theodoros H. and Zarouchas, Dimitrios}, 
    year={2023},
    OPTpages={105502} 
}

@article{Ofir_2025,
    title={{Real-Time Damage Detection in an Airplane Wing During Wind Tunnel Testing Under Realistic Flight Conditions}}, 
    OPTvolume={25}, 
    OPTISSN={1424-8220}, 
    DOI={10.3390/s25144423}, 
    OPTnumber={14}, 
    journal={Sensors}, 
    publisher={{MDPI AG}}, 
    author={Ofir, Yoav and Ben-Simon, Uri and Shoham, Shay and Kressel, Iddo and Galasso, Bernardino and Mercurio, Umberto and Concilio, Antonio and Apuleo, Gianvito and Bohbot, Jonathan and Tur, Moshe}, 
    year={2025},
    OPTpages={4423} 
}

@article{Broer_2021, 
    title={Fusion-based damage diagnostics for stiffened composite panels}, 
    OPTvolume={21}, 
    OPTISSN={1741-3168},
    DOI={10.1177/14759217211007127}, 
    OPTnumber={2}, 
    journal={{Structural Health Monitoring}}, 
    publisher={{SAGE Publications}}, 
    author={Broer, Agnes and Galanopoulos, Georgios and Benedictus, Rinze and Loutas, Theodoros and Zarouchas, Dimitrios}, 
    year={2021}, 
    OPTpages={613–639} 
}

@article{Tofallis_2015,
    title={A better measure of relative prediction accuracy for model selection and model estimation}, 
    OPTvolume={66}, 
    OPTISSN={1476-9360}, 
    DOI={10.1057/jors.2014.103}, 
    OPTnumber={8}, 
    journal={{Journal of the Operational Research Society}}, 
    publisher={{Informa UK Limited}}, 
    author={Tofallis, Chris}, 
    year={2015}, 
    OPTpages={1352–1362} 
}

@article{Kim_2016, 
    title={A new metric of absolute percentage error for intermittent demand forecasts}, 
    OPTvolume={32}, 
    OPTISSN={0169-2070}, 
    DOI={10.1016/j.ijforecast.2015.12.003},
    OPTnumber={3}, 
    journal={{International Journal of Forecasting}}, 
    publisher={{Elsevier BV}}, 
    author={Kim, Sungil and Kim, Heeyoung}, 
    year={2016}, 
    OPTpages={669–679} 
}

@article{Wang_2024, 
    title={{Structural Dynamic Response Reconstruction Based on Recurrent Neural Network–Aided Kalman Filter}}, 
    OPTvolume={2024}, 
    OPTISSN={1545-2263}, 
    DOI={10.1155/2024/7481513}, 
    OPTnumber={1}, 
    journal={{Structural Control and Health Monitoring}}, 
    publisher={Wiley}, 
    author={Wang, Yiqing and Song, Mingming and Wang, Ao and Sun, Limin}, 
    OPTeditor={Shi, Xiang}, 
    year={2024}
}

@article{Zheng_2025, 
    title={Data augmentation of dynamic responses for structural health monitoring using denoising diffusion probabilistic models}, 
    OPTvolume={328}, 
    OPTISSN={0141-0296}, 
    OPTDOI={10.1016/j.engstruct.2025.119685}, 
    journal={Engineering Structures}, 
    publisher={{Elsevier BV}}, 
    author={Zheng, Wenhao and Li, Jun and Hao, Hong}, 
    year={2025}, 
    OPTpages={119685} 
}

@article{Miele_2023, 
    title={Multi-fidelity physics-informed machine learning for probabilistic damage diagnosis}, 
    OPTvolume={235}, 
    OPTISSN={0951-8320},
    DOI={10.1016/j.ress.2023.109243}, 
    journal={{Reliability Engineering \& System Safety}}, 
    publisher={{Elsevier BV}}, 
    author={Miele, S. and Karve, P. and Mahadevan, S.}, 
    year={2023}, 
    OPTpages={109243} 
}

@article{Dafydd_2019, 
    title={Analysis of barely visible impact damage severity with ultrasonic guided Lamb waves}, 
    OPTvolume={19}, 
    OPTISSN={1741-3168}, 
    DOI={10.1177/1475921719878850}, 
    OPTnumber={4}, 
    journal={Structural Health Monitoring}, 
    publisher={{SAGE} Publications}, 
    author={Dafydd, Ifan and Sharif Khodaei, Zahra},
    year={2019},
    OPTpages={1104–1122} 
}

@incollection{Giurgiutiu_2020, 
    title={{Structural Health Monitoring (SHM)} of aerospace composites}, 
    ISBN={9780081026793}, 
    DOI={10.1016/b978-0-08-102679-3.00017-4}, 
    booktitle={{Polymer Composites in the Aerospace Industry}}, 
    publisher={Elsevier},
    author={Giurgiutiu, Victor}, 
    year={2020}, 
    pages={491–558} 
}

@article{Seno_2020,
    title={A novel method for impact force estimation in composite plates under simulated environmental and operational conditions}, 
    OPTvolume={29}, 
    OPTISSN={1361-665X}, 
    DOI={10.1088/1361-665x/abb06e}, 
    OPTnumber={11}, journal={Smart Materials and Structures}, 
    publisher={{IOP} Publishing}, 
    author={Seno, Aldyandra Hami and Aliabadi, M H Ferri}, 
    year={2020},  
    OPTpages={115029} 
}

@article{Seno_2021, 
    title={Uncertainty quantification for impact location and force estimation in composite structures}, 
    OPTvolume={21},
    OPTISSN={1741-3168}, 
    DOI={10.1177/14759217211020255}, 
    OPTnumber={3},
    journal={{Structural Health Monitoring}}, 
    publisher={{SAGE} Publications},
    author={Seno, Aldyandra Hami and Aliabadi, MH Ferri}, 
    year={2021},
    OPTpages={1061–1075} 
}

@article{Datta_2019, 
    title={{Impact Localization and Severity Estimation on Composite Structure Using Fiber Bragg Grating Sensors by Least Square Support Vector Regression}}, 
    OPTvolume={19}, 
    OPTISSN={2379-9153},
    DOI={10.1109/jsen.2019.2901453}, 
    OPTnumber={12}, 
    journal={{IEEE Sensors Journal}}, 
    publisher={Institute of Electrical and Electronics Engineers {(IEEE)}}, 
    author={Datta, Amitabha and Augustin, M. J. and Gupta, Nitesh and Viswamurthy, S. R. and Gaddikeri, Kotresh M. and Sundaram, Ramesh}, 
    year={2019},
    OPTpages={4463–4470} 
}

@article{Zhou_2019, 
    title={Finite element analysis of a modified progressive damage model for composite laminates under low-velocity impact},
    volume={225}, 
    ISSN={0263-8223}, 
    DOI={10.1016/j.compstruct.2019.111113}, 
    journal={Composite Structures}, 
    publisher={Elsevier}, 
    author={Zhou, Junjie and Wen, Pihua and Wang, Shengnan}, 
    year={2019},
    pages={111113} 
}

@article{Molina_Viedma_2021, 
    title={{Full-Field Operational Modal Analysis of an Aircraft Composite Panel from the Dynamic Response in Multi-Impact Test}}, 
    OPTvolume={21},
    OPTISSN={1424-8220},
    DOI={10.3390/s21051602}, 
    OPTnumber={5}, 
    journal={Sensors}, 
    publisher={{MDPI AG}}, 
    author={{Molina-Viedma, Ángel and L\'{o}pez-Alba, El\'{i}as and Felipe-Ses\'{e}, Luis and D\'{i}az, Francisco}}, 
    year={2021},
    OPTpages={1602}
}

@article{Ooijevaar_2015,
    title={Impact damage identification in composite skin-stiffener structures based on modal curvatures: Damage Identification in Skin-stiffener Structures Based on Curvatures}, 
    OPTvolume={23}, 
    OPTISSN={1545-2255}, 
    DOI={10.1002/stc.1754}, 
    OPTnumber={2}, 
    journal={Structural Control and Health Monitoring}, 
    publisher={Wiley}, 
    author={Ooijevaar, T.H. and Warnet, L.L. and Loendersloot, R. and Akkerman, R. and Tinga, T.},
    year={2015},
    OPTpages={198–217} 
}

@article{Stephen_2021, 
    title={Energy absorption and damage assessment of non-hybrid and hybrid fabric epoxy composite laminates: experimental and numerical study}, 
    OPTvolume={14},
    OPTISSN={2238-7854}, 
    DOI={10.1016/j.jmrt.2021.08.108}, 
    journal={{Journal of Materials Research and Technology}},
    publisher={Elsevier}, 
    author={Stephen, Clifton and Mourad, Abdel-Hamid. I. and Shivamurthy, B. and Selvam, Rajiv},
    year={2021}, 
    OPTpages={3080–3091} 
}

@article{Sellami_2025,
    title={{Guided Wave Propagation in a Realistic CFRP Fuselage Panel: Proof of Concept for Early-Stage Damage Detection}}, 
    OPTvolume={25}, 
    OPTISSN={1424-8220},
    DOI={10.3390/s25041104}, 
    OPTnumber={4}, 
    journal={Sensors}, 
    publisher={{MDPI AG}}, 
    author={Sellami, Fatma and Memmolo, Vittorio and Hornung, Mirko},
    year={2025}, 
    pages={1104} 
}

@article{Zhang_2017,
    title={An integrated numerical model for investigating guided waves in impact-damaged composite laminates},
    OPTvolume={176},
    OPTISSN={0263-8223}, 
    DOI={10.1016/j.compstruct.2017.06.034}, 
    journal={Composite Structures}, 
    publisher={Elsevier}, 
    author={Zhang, B. and Sun, X.C. and Eaton, M.J. and Marks, R. and Clarke, A. and Featherston, C.A. and Kawashita, L.F. and Hallett, S.R.}, 
    year={2017},
    OPTpages={945–960} 
}

@article{Bogenfeld_2018, 
    title={Review and benchmark study on the analysis of low-velocity impact on composite laminates}, 
    OPTvolume={86},
    OPTISSN={1350-6307}, 
    DOI={10.1016/j.engfailanal.2017.12.019}, 
    journal={{Engineering Failure Analysis}},
    publisher={Elsevier},
    author={Bogenfeld, Raffael and Kreikemeier, Janko and Wille, Tobias},
    year={2018}, 
    OPTpages={72–99} 
}

@article{Alonso_2021, 
    title={{High-velocity impact on composite sandwich structures: A theoretical model}},
    OPTvolume={201}, 
    OPTISSN={0020-7403},
    DOI={10.1016/j.ijmecsci.2021.106459}, 
    journal={International Journal of Mechanical Sciences}, 
    publisher={Elsevier}, 
    author={Alonso, L. and Solis, A.}, 
    year={2021}, 
    OPTpages={106459} 
}

@article{Zhu_2020,
    title={Dynamic response of foam core sandwich panel with composite facesheets during low-velocity impact and penetration}, 
    OPTvolume={139}, 
    OPTISSN={0734-743X},
    DOI={10.1016/j.ijimpeng.2020.103508}, 
    journal={{International Journal of Impact Engineering}}, 
    publisher={Elsevier}, 
    author={Zhu, Yefei and Sun, Yuguo}, 
    year={2020}, 
    OPTpages={103508}
}

@article{Shariyat_2018,
    title={A new analytical solution and novel energy formulations for non-linear eccentric impact analysis of composite multi-layer/sandwich plates resting on point supports}, 
    OPTvolume={127}, 
    OPTISSN={0263-8231}, 
    DOI={10.1016/j.tws.2018.02.001}, 
    journal={Thin-Walled Structures}, 
    publisher={Elsevier},
    author={Shariyat, M. and Roshanfar, M.}, 
    year={2018},
    pages={157–168} 
}

@article{Jung_2021, 
    title={Advanced deep learning model-based impact characterization method for composite laminates}, 
    OPTvolume={207}, 
    OPTISSN={0266-3538},
    DOI={10.1016/j.compscitech.2021.108713}, 
    journal={{Composites Science and Technology}}, 
    publisher={Elsevier}, 
    author={Jung, Kyung-Chae and Chang, Seung-Hwan},
    year={2021}, 
    pages={108713}
}

@article{Zhu_2023, 
    title={Impact energy assessment of sandwich composites using an ensemble approach boosted by deep learning and electromechanical impedance},
    OPTvolume={32}, 
    ISSN={1361-665X}, 
    DOI={10.1088/1361-665x/ace868},
    OPTnumber={9},
    journal={{Smart Materials and Structures}}, 
    publisher={{IOP Publishing}}, 
    author={Zhu, Jianjian and Wen, Jinshan and Han, Zhibin and Ho, Mabel Mei-po and Lan, Zifeng and Wang, Yishou and Qing, Xinlin},
    year={2023},
    OPTpages={095019}
}

@article{Zhong_2015, 
    title={Impact energy level assessment of composite structures using {MUSIC-ANN} approach: {MUSIC-ANN} approach-based impact monitoring for composite structures}, 
    OPTvolume={23},
    OPTISSN={1545-2255},
    DOI={10.1002/stc.1815},
    OPTnumber={5},
    journal={{Structural Control and Health Monitoring}}, 
    publisher={Hindawi Limited}, 
    author={Zhong, Yongteng and Xiang, Jiawei and Gao, Haifeng and Zhou, Yuqing}, year={2015},
    OPTpages={825–837}
}

@article{Xiao_2025, 
    title={Robust impact localisation on composite aerostructures using kernel design and Bayesian-inspired model averaging under environmental and operational uncertainties}, 
    OPTISSN={1741-3168}, 
    DOI={10.1177/14759217251362397}, 
    journal={{Structural Health Monitoring}},
    publisher={{SAGE Publications}}, 
    author={Xiao, Dong and Sharif-Khodaei, Zahra and Aliabadi, MH}, 
    year={2025}
}

@article{Rautela_2021, 
    title={Combined two-level damage identification strategy using ultrasonic guided waves and physical knowledge assisted machine learning}, 
    OPTvolume={115}, 
    OPTISSN={0041-624X}, 
    DOI={10.1016/j.ultras.2021.106451}, 
    journal={Ultrasonics}, 
    publisher={Elsevier BV},
    author={Rautela, Mahindra and Senthilnath, J. and Moll, Jochen and Gopalakrishnan, Srinivasan},
    year={2021}, 
    OPTpages={106451}
}

@article{Torzoni_2023,
    title={A multi-fidelity surrogate model for structural health monitoring exploiting model order reduction and artificial neural networks},
    OPTvolume={197}, 
    OPTISSN={0888-3270}, 
    DOI={10.1016/j.ymssp.2023.110376}, 
    journal={Mechanical Systems and Signal Processing}, 
    publisher={Elsevier}, 
    author={Torzoni, Matteo and Manzoni, Andrea and Mariani, Stefano}, 
    year={2023}, 
    pages={110376}
}

@article{Zou_2023, 
    title={Virtual sensing of subsoil strain response in monopile-based offshore wind turbines via Gaussian process latent force models},
    OPTvolume={200}, 
    OPTISSN={0888-3270}, 
    DOI={10.1016/j.ymssp.2023.110488}, 
    journal={Mechanical Systems and Signal Processing}, 
    publisher={Elsevier}, 
    author={Zou, Joanna and Lourens, Eliz-Mari and Cicirello, Alice}, 
    year={2023},
    OPTpages={110488} 
}

@article{Malekloo_2021, 
    title={Machine learning and structural health monitoring overview with emerging technology and high-dimensional data source highlights}, 
    OPTvolume={21}, 
    OPTISSN={1741-3168}, 
    DOI={10.1177/14759217211036880}, 
    OPTnumber={4}, 
    journal={Structural Health Monitoring}, 
    publisher={{SAGE} Publications}, 
    author={Malekloo, Arman and Ozer, Ekin and AlHamaydeh, Mohammad and Girolami, Mark}, 
    year={2021}, 
    OPTpages={1906–1955} 
}

@article{Jia_2023, 
     title={{Deep Learning for Structural Health Monitoring: Data, Algorithms, Applications, Challenges, and Trends}},
     OPTvolume={23}, 
     OPTISSN={1424-8220},
     DOI={10.3390/s23218824}, 
     OPTnumber={21}, 
     journal={Sensors}, 
     publisher={{MDPI AG}},
     author={Jia, Jing and Li, Ying}, 
     year={2023}, 
     OPTpages={8824} 
 }

@article{Azimi_2020,
     title={Data-Driven Structural Health Monitoring and Damage Detection through Deep Learning: State-of-the-Art Review}, 
     OPTvolume={20}, 
     OPTISSN={1424-8220}, 
     DOI={10.3390/s20102778}, 
     OPTnumber={10}, 
     journal={Sensors}, 
     publisher={{MDPI AG}}, 
     author={Azimi, Mohsen and Eslamlou, Armin and Pekcan, Gokhan}, 
     year={2020}, 
     OPTpages={2778} 
 }

@article{Nicassio_2025, 
    title={{Energy Evaluation and Passive Damage Detection for Structural Health Monitoring in Aerospace Structures Using Machine Learning Models}}, 
    OPTvolume={25}, 
    OPTISSN={1424-8220},
    DOI={10.3390/s25164942}, 
    OPTnumber={16}, 
    journal={Sensors}, 
    publisher={{MDPI AG}}, 
    author={Nicassio, Francesco and Dipietrangelo, Flavio and Gaspari, Antonella and Scarselli, Gennaro}, 
    year={2025}, 
    OPTpages={4942} 
}

@article{Kralovec_2020, 
    title={{Review of Structural Health Monitoring Methods Regarding a Multi-Sensor Approach for Damage Assessment of Metal and Composite Structures}}, 
    OPTvolume={20}, 
    OPTISSN={1424-8220}, 
    DOI={10.3390/s20030826}, 
    OPTnumber={3}, 
    journal={Sensors}, 
    publisher={{MDPI AG}}, 
    author={Kralovec, Christoph and Schagerl, Martin}, 
    year={2020},
    OPTpages={826}
}

@article{Fan_2021, 
    title={Review of piezoelectric impedance based structural health monitoring: Physics-based and data-driven methods}, 
    OPTvolume={24}, 
    OPTISSN={2048-4011},
    DOI={10.1177/13694332211038444}, 
    OPTnumber={16}, 
    journal={{Advances in Structural Engineering}}, 
    publisher={{SAGE Publications}},
    author={Fan, Xingyu and Li, Jun and Hao, Hong}, 
    year={2021}, 
    OPTpages={3609–3626} 
}

@article{Hassani_2021, 
    title={{Structural Health Monitoring in Composite Structures: A Comprehensive Review}}, 
    OPTvolume={22}, 
    OPTISSN={1424-8220},
    DOI={10.3390/s22010153},
    OPTnumber={1}, 
    journal={Sensors},
    publisher={{MDPI AG}}, 
    author={Hassani, Sahar and Mousavi, Mohsen and Gandomi, Amir H.}, 
    year={2021},
    OPTpages={153} 
}

@article{Blakseth_2022,
    title={Combining physics-based and data-driven techniques for reliable hybrid analysis and modeling using the corrective source term approach}, 
    OPTvolume={128}, 
    OPTISSN={1568-4946}, 
    DOI={10.1016/j.asoc.2022.109533}, 
    journal={Applied Soft Computing}, 
    publisher={Elsevier}, 
    author={Blakseth, Sindre Stenen and Rasheed, Adil and Kvamsdal, Trond and San, Omer}, 
    year={2022}, 
    OPTpages={109533} 
}

@article{Arias_Chao_2022,
    title={Fusing physics-based and deep learning models for prognostics}, 
    OPTvolume={217}, 
    OPTISSN={0951-8320}, 
    DOI={10.1016/j.ress.2021.107961}, 
    journal={Reliability Engineering \& System Safety},
    publisher={Elsevier}, 
    author={Arias Chao, Manuel and Kulkarni, Chetan and Goebel, Kai and Fink, Olga},
    year={2022}, 
    OPTpages={107961} 
}

@article{Ni_2020, 
    title={A Bayesian approach for condition assessment and damage alarm of bridge expansion joints using long-term structural health monitoring data}, 
    OPTvolume={212},
    OPTISSN={0141-0296}, 
    OPTDOI={10.1016/j.engstruct.2020.110520}, 
    journal={Engineering Structures}, 
    publisher={Elsevier}, 
    author={Ni, Y.Q. and Wang, Y.W. and Zhang, C.},
    year={2020}, 
    OPTpages={110520}
}

@article{Wang_2025, 
    title={{Bayesian Network in Structural Health Monitoring: Theoretical Background and Applications Review}}, 
    OPTvolume={25}, 
    OPTISSN={1424-8220}, 
    DOI={10.3390/s25123577},
    OPTnumber={12}, 
    journal={Sensors},
    publisher={{MDPI AG}}, 
    author={Wang, Qi-Ang and Lu, Ao-Wen and Ni, Yi-Qing and Wang, Jun-Fang and Ma, Zhan-Guo}, 
    year={2025}, 
    OPTpages={3577} 
}

@article{Hughes_2021, 
    title={A probabilistic risk-based decision framework for structural health monitoring}, 
    OPTvolume={150},
    OPTISSN={0888-3270}, 
    DOI={10.1016/j.ymssp.2020.107339}, 
    journal={Mechanical Systems and Signal Processing},
    publisher={Elsevier}, 
    author={Hughes, A.J. and Barthorpe, R.J. and Dervilis, N. and Farrar, C.R. and Worden, K.}, 
    year={2021}, 
    OPTpages={107339} 
}

@article{Cuomo_2022, 
     title={Scientific Machine Learning Through Physics–Informed Neural Networks: Where we are and What’s Next}, 
     OPTvolume={92}, 
     OPTISSN={1573-7691},
     DOI={10.1007/s10915-022-01939-z}, 
     OPTnumber={3}, 
     journal={Journal of Scientific Computing},
     publisher={{Springer Science and Business Media LLC}},
     author={Cuomo, Salvatore and Di Cola, Vincenzo Schiano and Giampaolo, Fabio and Rozza, Gianluigi and Raissi, Maziar and Piccialli, Francesco},
     year={2022}
 }

@article{Xu_2023, 
    title={Physics-informed machine learning for reliability and systems safety applications: State of the art and challenges}, 
    OPTvolume={230},
    OPTISSN={0951-8320}, 
    DOI={10.1016/j.ress.2022.108900}, 
    journal={Reliability Engineering \& System Safety}, 
    publisher={Elsevier}, 
    author={Xu, Yanwen and Kohtz, Sara and Boakye, Jessica and Gardoni, Paolo and Wang, Pingfeng}, 
    year={2023},
    OPTpages={108900} 
}

@article{Wu_2024, 
    title={Physics-informed machine learning: A comprehensive review on applications in anomaly detection and condition monitoring}, 
    OPTvolume={255}, 
    OPTISSN={0957-4174}, 
    DOI={10.1016/j.eswa.2024.124678}, 
    journal={Expert Systems with Applications}, 
    publisher={Elsevier}, 
    author={Wu, Yuandi and Sicard, Brett and Gadsden, Stephen Andrew},
    year={2024}, 
    OPTpages={124678}
}

\end{document}